\documentclass[11pt]{article}

\usepackage[final]{acl}

\usepackage{times}
\usepackage{latexsym}
\usepackage{enumitem}
\usepackage{makecell}

\usepackage[T1]{fontenc}

\usepackage[utf8]{inputenc}

\usepackage{microtype}

\usepackage{inconsolata}

\usepackage{graphicx}

\usepackage{amsmath}
\usepackage{pifont}
\newcommand{\cmark}{\ding{51}}   
\newcommand{\xmark}{\ding{55}}   
\usepackage{booktabs}   
\usepackage{siunitx}    
\usepackage{caption}    
\usepackage{multirow}
\usepackage{listings}
\usepackage{xcolor}
\usepackage{subcaption}

\title{DMT-CBT: Longitudinal Therapeutic State Modeling for CBT Counseling}

\author{
 \textbf{Chang Liu\textsuperscript{1}},
 \textbf{Shuyi Zhang\textsuperscript{1}},
 \textbf{Changsheng Ma\textsuperscript{1}},
 \textbf{Yongfeng Tao\textsuperscript{1}},
 \textbf{Minqiang Yang\textsuperscript{1}},
\\
 \textbf{Bin Hu\textsuperscript{1}},
\\
\\
\\
 \textsuperscript{1}School of Information Science and Engineering, Lanzhou University,
\\
}

\begin{document}
\maketitle
\begin{abstract}
Large language models (LLMs) have shown growing potential for Cognitive Behavioral Therapy (CBT) counseling. However, most existing approaches still formulate counseling as a local response generation problem, focusing on empathetic replies within short, text-only, or single-session interactions. We argue that this formulation fundamentally mismatches the nature of real psychotherapy. In clinical CBT, therapy is a longitudinal process in which therapists continuously infer, update, and intervene on evolving therapeutic states across sessions. Realistic CBT further involves multimodal inference and delayed cross-session intervention effects, requiring models to capture longitudinal therapeutic state evolution under partial observability.
We propose DMT-CBT, a framework for \underline{D}ynamic \underline{M}odeling of evolving \underline{T}herapeutic states in CBT counseling. DMT-CBT maintains structured therapeutic states across sessions while incorporating multimodal behavioral grounding and tool-augmented intervention to support adaptive therapeutic reasoning. Based on this framework, we construct DMTCorpus, a synthetic multi-session multimodal CBT counseling dataset featuring evolving therapeutic states, image-grounded client behaviors, and cross-session intervention continuity.
Experimental results show that DMT-CBT improves counseling fidelity and therapeutic alliance, produces more favorable longitudinal affective trajectories, and preserves therapeutic states more faithfully than post-hoc extraction approaches.

\end{abstract}

\section{Introduction}

Mental health disorders remain one of the most widespread yet insufficiently addressed global health challenges, with many individuals facing barriers to psychological care due to workforce shortages, financial burden, limited accessibility, and persistent social stigma \citep{world2022world, world2025over}. Recent advances in large language models (LLMs) have created new opportunities for scalable mental health support, particularly for Cognitive Behavioral Therapy (CBT), whose structured and evidence-based treatment process naturally aligns with language-based interaction~\citep{shen2024large}.

\begin{figure}[t]
  \includegraphics[width=\columnwidth]{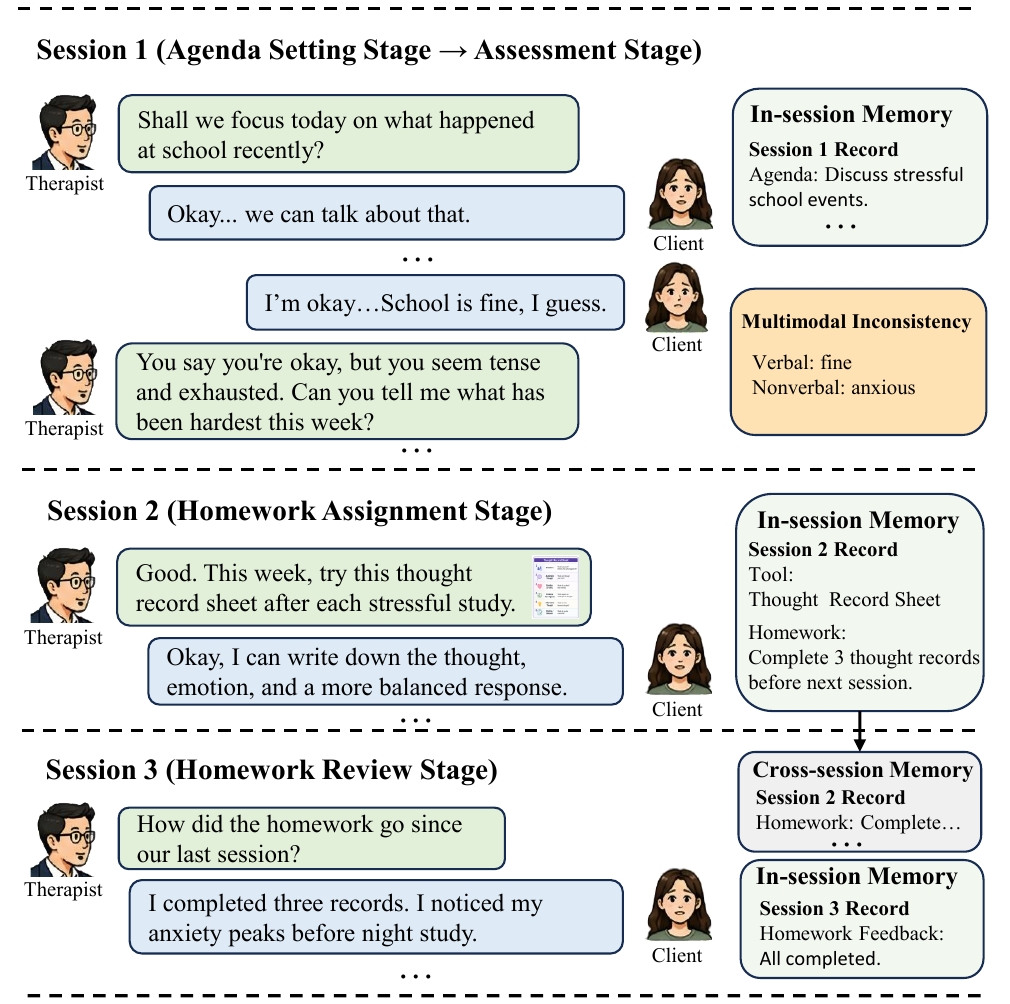}
  \caption{A motivating example of a realistic multi-session CBT counseling process.}
  \label{fig:intro}
\end{figure}

Recent LLM-based counseling systems have begun incorporating CBT principles into therapeutic dialogue generation. However, most existing approaches still implicitly formulate counseling as a local response generation problem, focusing on empathetic replies within short, text-only, or single-session interactions~\citep{na-2024-cbt,lee2024cactus}. We argue that this formulation fundamentally mismatches the nature of real psychotherapy. In clinical CBT, therapy is a longitudinal process in which therapists continuously infer, update, and intervene on evolving therapeutic states across sessions. Effective treatment therefore depends on modeling therapeutic progression over time rather than optimizing isolated responses alone.

Figure~\ref{fig:intro} illustrates why realistic CBT cannot be reduced to response generation. Therapists often infer latent client conditions from multimodal behavioral cues, particularly when verbal and nonverbal signals are inconsistent. Moreover, CBT interventions frequently exhibit delayed cross-session effects, where therapists revisit prior homework and adapt future interventions based on the client’s longitudinal progression. Consequently, realistic CBT requires modeling longitudinal therapeutic state evolution under partial observability.

Although recent work has explored multimodal counseling and multi-session dialogue generation, existing systems still fail to model longitudinal therapeutic state evolution under realistic interaction constraints. Multimodal approaches often treat visual information as static auxiliary input rather than behavioral evidence for therapeutic inference~\citep{kim2025multimodal,kim2025mirror}. Multi-session systems typically rely on dialogue history or coarse summaries instead of structured therapeutic states that support adaptive reasoning across sessions~\citep{zhou2025diacbt,wang2025psychological,pan2026psycheval}. Moreover, many simulation frameworks adopt an omniscient generation paradigm, where a single model directly accesses both client and therapist states, failing to reflect the partial observability inherent in real psychotherapy~\citep{lee2024cactus,xie2025psydt}.

To address these limitations, we propose DMT-CBT, a dynamic multimodal framework for longitudinal CBT process modeling. DMT-CBT maintains structured therapeutic states across sessions to support persistent tracking of cognitive beliefs, emotional trajectories, intervention progress, and homework continuity. The framework further incorporates multimodal behavioral grounding and tool-augmented intervention to enable adaptive therapeutic reasoning under partial observability.

Based on this framework, we construct DMTCorpus, a synthetic multi-session multimodal CBT counseling dataset designed to capture longitudinal therapeutic progression. We evaluate DMT-CBT from three complementary perspectives: session-level counseling quality, longitudinal affective trajectories, and system-level module reliability. Experimental results show that DMT-CBT improves counseling fidelity and therapeutic alliance, produces more favorable affective trajectories across sessions, and preserves therapeutic states more faithfully than post-hoc extraction approaches.

Our contributions are summarized as follows:
\begin{itemize}[nosep]
\item We reformulate LLM-based CBT counseling as a longitudinal therapeutic state modeling problem rather than isolated empathetic response generation.

\item We propose DMT-CBT, a multimodal and tool-augmented framework for modeling therapeutic state evolution and adaptive intervention under partial observability.

\item We construct DMTCorpus, a synthetic multi-session multimodal CBT dataset featuring evolving therapeutic states, image-grounded client behaviors, and cross-session intervention continuity.

\item We conduct multi-level evaluations covering counseling quality, longitudinal affective trajectories, and system reliability, demonstrating the effectiveness of DMT-CBT for realistic CBT process modeling.

\end{itemize}

\begin{figure*}[t]
  \includegraphics[width=0.95\linewidth] {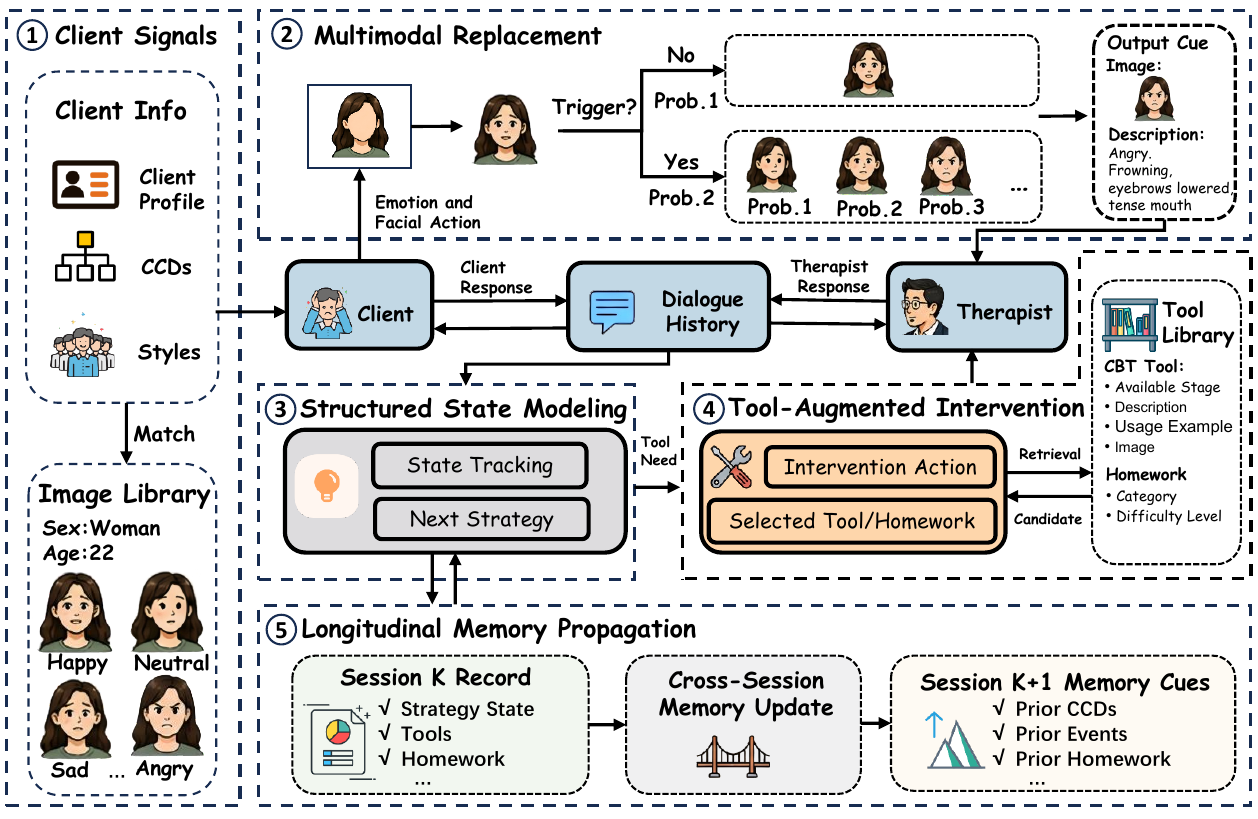} 
 
  \caption {Overview of the DMT-CBT framework.}
    \label{fig:framework}
\end{figure*}

\section{Related Work}

\paragraph{LLM-based CBT Counseling.}
Recent studies have explored LLM-based CBT counseling by incorporating therapeutic strategies, role-playing mechanisms, or empathetic response generation into dialogue systems~\citep{na-2024-cbt,lee2024cactus,xie2025psydt,anonymous2026ccd}. However, most existing approaches still formulate counseling as a local response generation problem, focusing on producing appropriate therapist replies within short or isolated interactions. In contrast, we model CBT counseling as longitudinal therapeutic state evolution across sessions.

\paragraph{Longitudinal Therapeutic Process Modeling.}
Recent work has extended counseling systems from single-session interaction toward multi-session therapeutic simulation~\citep{zhou2025diacbt,wang2025psychological,pan2026psycheval}. Related studies on memory-augmented agents also explore long-term conversational memory and user modeling~\citep{park2023generative,zhong2024memorybank,li2025toward,wang2025annaagent}. However, existing approaches primarily rely on dialogue history or coarse summaries~\citep{hu2026theramind}, which are insufficient for modeling structured therapeutic progression over time. Moreover, many frameworks adopt omniscient interaction settings that fail to reflect the partial observability inherent in real psychotherapy~\citep{zhou2025diacbt,pan2026psycheval}.

\paragraph{Multimodal Therapeutic Interaction.}
Recent multimodal counseling systems incorporate visual information into therapeutic dialogue and emotional understanding~\citep{kim2025multimodal,kim2025mirror}. Related work in affective computing further highlights the importance of nonverbal behavioral signals for emotion understanding~\citep{poria2017review,tsai2019multimodal}. Nevertheless, existing approaches often treat multimodal signals as auxiliary perceptual inputs rather than behavioral evidence for therapeutic inference. Our work instead incorporates multimodal behavioral grounding into longitudinal therapeutic state modeling.

\section{Method}\label{method}

\subsection{Problem Formulation}

We formulate CBT counseling as a longitudinal therapeutic process in which therapists continuously infer and intervene on evolving client states across sessions. Unlike conventional response-generation settings, psychotherapy requires maintaining therapeutic progression from partially observable client behaviors and delayed intervention effects over time.

Given a counseling case consisting of multiple sessions, the goal is to model the longitudinal evolution of therapeutic states throughout counseling. At each dialogue turn $t$, the therapist observes client-side signals and generates therapeutic interventions conditioned on the current counseling context. Formally, we represent the counseling trajectory as
\begin{equation}
\mathcal{T} = \{(o_t, h_t, u_t)\}_{t=1}^{T},
\end{equation}
where $o_t$ denotes observable client-side signals, $h_t$ denotes the therapist-side counseling state, and $u_t$ denotes the therapist intervention at turn $t$. To operationalize this process, we propose DMT-CBT, as illustrated in Figure~\ref{fig:framework}, which integrates structured counseling-state tracking, multimodal behavioral grounding, cross-session memory propagation, and tool-augmented intervention within a unified longitudinal counseling framework.

\subsection{Therapeutic State Modeling}

DMT-CBT maintains structured counseling states to model longitudinal CBT progression beyond raw dialogue history or free-form summaries. For client $i$ in session $k$, we define a session-dependent therapeutic condition
\begin{equation}
c_i^{(k)} =
\begin{cases}
\text{profile}_i, & k = 1, \\
\mathrm{CCD}^{\text{client}}_{i,k}, & k > 1,
\end{cases}
\end{equation}
where the initial session is grounded in the client profile and later sessions are conditioned on evolving CBT conceptualization records.

During interaction, the framework maintains a therapist-side counseling state $h_t$ through stage-aware therapeutic reasoning:
\begin{equation}
z_t,\, \Delta h_t
=
\mathcal{LLM}(q_t,\, h_t,\, o_t),
\end{equation}
where $q_t=(p_t,g_t,p_{t+1})$ denotes the current therapeutic stage, stage-specific guidance, and expected next-stage transition. The maintained counseling state primarily captures CBT-relevant cognitive patterns, emotional conditions, behavioral tendencies, and intervention progress throughout counseling.

The output $z_t$ predicts whether the current therapeutic objective is completed, while $\Delta h_t$ denotes newly extracted counseling-state updates. The counseling state is incrementally updated as
\begin{equation}
h_{t+1} = \Phi(h_t,\Delta h_t),
\end{equation}
where $\Phi(\cdot)$ denotes a structured update function.

\subsection{Longitudinal Therapeutic Evolution}

Beyond in-session interaction, CBT counseling requires maintaining therapeutic continuity across sessions. At the end of session $k$, the finalized counseling state is stored as a structured counseling record $R_i^{(k)}$. During session $k+1$, the framework retrieves stage-aware therapeutic cues conditioned on both prior counseling records and the predefined CBT treatment progression:
\begin{equation}
\tilde{h}^{(k+1)}_t
=
\Psi(R_i^{(k)},\, \mathcal{P}^{\text{CBT}},\, p_t),
\end{equation}
where $\mathcal{P}^{\text{CBT}}$ denotes the multi-session CBT progression structure and $\Psi(\cdot)$ denotes a stage-aware memory retrieval function.

The retrieved cues support cross-session therapeutic continuity, including homework revisiting, delayed intervention tracking, and adaptive strategy adjustment. This forms a hierarchical memory mechanism consisting of incremental within-session state tracking and cross-session state propagation.

\subsection{Multimodal Grounding}

In psychotherapy, client verbal responses may not fully reflect latent psychological conditions. DMT-CBT therefore models observable client behavior as
\begin{equation}
o_t = (r_t, m_t),
\end{equation}
where $r_t$ denotes the client’s verbal response and $m_t=(e_t,a_t)$ denotes structured multimodal behavioral observations, including emotion labels and behavioral-action descriptions.

Client behaviors are generated conditioned on the evolving therapeutic condition:
\begin{equation}
r_t,\, e_t,\, a_t
=
\mathcal{LLM}(c_i^{(k)},\, s_i,\, \mathcal{H}_{t-1},\, \tau_t),
\end{equation}
where $s_i$ denotes one of six counseling-relevant interpersonal styles (Appendix~\ref{app:styles}) and $\tau_t$ denotes the currently activated intervention tool when applicable.

Importantly, DMT-CBT explicitly models verbal--nonverbal inconsistency during multimodal grounding. Behavioral cues may not always align with literal client utterances, reflecting realistic psychotherapy interactions. Structured behavioral observations are further mapped to facial-expression images using the constructed facial-expression library (Appendix~\ref{app:replacement}).

\subsection{Tool-Augmented Therapeutic Intervention}

CBT counseling frequently relies on structured intervention tools and homework assignments beyond verbal interaction alone. For in-session intervention, DMT-CBT retrieves candidate CBT tools from a visual tool library using a dense retriever based on BGE-large-zh-v1.5~\citep{cui-etal-2020-revisiting}. Conditioned on the therapeutic stage, dialogue context, and intervention history, the framework predicts a tool decision:
\begin{equation}
\omega_t
=
\mathcal{LLM}(p_t,\, p_{t+1},\, \tau_t,\, \mathcal{C}^{\text{tool}}_t,\, \mathcal{H}_{t-1}),
\end{equation}
where $\omega_t$ specifies both the intervention action and the selected CBT tool. Detailed retrieval settings are provided in Appendix~\ref{app:mm_tool}.

For cross-session continuity, the framework additionally generates structured homework assignments:
\begin{equation}
\eta_t
=
\mathcal{LLM}(h_t,\, \mathcal{C}^{\text{hw}}_t),
\end{equation}
where $\eta_t$ denotes the recommended homework intervention.

\begin{table*}[t]
  \centering
  \small
  \setlength{\tabcolsep}{6pt}
  \renewcommand{\arraystretch}{1.12}
  \begin{tabular}{lccccccc}
    \toprule
    \textbf{Datasets} & \textbf{Modality} & \textbf{Cases} & \textbf{Avg. Sess.} & \textbf{Avg. Turns} & \textbf{Total Sess.} & \textbf{Tool-Aug} & \textbf{Available} \\
    \midrule
    CBT-LLM~\citep{na-2024-cbt}            & T    & 22,327 & 1.0 & 1.0  & 22,327 & \xmark & \cmark \\
    HealMe~\citep{xiao-etal-2024-healme}   & T    & 1,300  & 1.0 & 3.0  & 1,300  & \xmark & \xmark \\
    CACTUS~\citep{lee2024cactus}           & T    & 31,577 & 1.0 & 16.6 & 31,577 & \xmark & \cmark \\
    PsyDTCorpus~\citep{xie2025psydt}       & T    & 4,760  & 1.0 & 18.0 & 4,760  & \xmark & \cmark \\
    M2CoSC~\citep{kim2025multimodal}                  & T, V & 429    & 1.0 & 4.0  & 429    & \xmark & \cmark \\
    MIRROR~\citep{kim2025mirror}                  & T, V & 3,073  & 1.0 & 10.3 & 3,073  & \xmark & \cmark \\
    Psy-Insight~\citep{chen2025psy}                   & T    & 189    & 5.0 & 6.5  & 951    & \xmark & \cmark \\
    PsychEval~\citep{pan2026psycheval}                     & T    & 369    & 7.6 & 24.1 & 2,798  & \xmark & \cmark \\
    \midrule
    \textbf{DMTCorpus}                     & \textbf{T, V} & \textbf{768} & \textbf{5.6} & \textbf{21.4} & \textbf{4,317} & \cmark & \cmark \\
    \bottomrule
  \end{tabular}
  \caption{\label{tab:dataset_comparison}
    Comparison of CBT-related counseling datasets. DMTCorpus is the only dataset in this comparison that simultaneously supports multimodality, multi-session interaction, and tool-augmented counseling. T denotes text and V denotes vision. Avg. Sess. indicates the average number of sessions per case, and Avg. Turns indicates the average number of turns per session. \cmark\ indicates support or availability; \xmark\ indicates not supported or unavailable.
  }
\end{table*}

\subsection{DMTCorpus Construction}

\paragraph{Data Resources.}
To instantiate DMT-CBT, we construct textual, multimodal, and intervention resources for longitudinal CBT simulation.

For textual resources, we collect 148 complete CBT counseling cases from PsychEval~\citep{pan2026psycheval}, each paired with client profiles and Cognitive Conceptualization Diagrams (CCDs)~\citep{wang-etal-2024-patient}. We further derive a generic multi-session CBT treatment structure from CBT manuals and counseling books~\citep{beck2011cognitive,beck2020cognitive}. In addition, we construct a homework repository containing 383 items across six general categories. More implementation details are provided in Appendix~\ref{app:text_resources}.

For multimodal grounding, we construct a facial-expression library based on DISFA~\citep{mavadati2013disfa}, containing 3,592 facial-expression images with structured action-unit annotations and emotion mappings~\citep{ekman1999basic}. We additionally build a visual CBT tool library containing worksheets, intervention diagrams, and homework-related materials collected from CBT manuals and counseling resources. More details are provided in Appendix~\ref{app:mm_resources}.

\paragraph{Counseling Simulation.}
Using the collected resources, we instantiate all functional components with GPT-4.1-mini. Specifically, we pair each counseling case with one of six predefined interpersonal styles to construct diverse client conditions (Appendix~\ref{app:styles}). For each condition, DMT-CBT generates a six-session counseling trajectory following the predefined CBT treatment progression, resulting in 768 case-style conditions and an initial pool of 4,608 sessions.

To further support multimodal counseling simulation, we apply a multimodal grounding strategy that maps generated emotion--action pairs to facial-expression images from the constructed image library. We additionally introduce controlled verbal--nonverbal incongruence, allowing behavioral cues to occasionally diverge from the literal semantic content of client utterances. For tool-grounded intervention, selected CBT tools are associated with corresponding visual intervention materials. Detailed replacement rules are provided in Appendix~\ref{app:replacement}.

\paragraph{Quality Control.}
We apply a three-stage quality-control pipeline to remove abnormal or clinically implausible sessions. First, we remove sessions with more than 30 dialogue turns or empty fields (\textbf{1.48\%} rejected). Second, we filter out sessions with excessively long or otherwise abnormal responses (\textbf{0.93\%} rejected). Finally, we evaluate counseling quality using CTRS-based filtering with GPT-4o-mini and remove sessions with obvious clinical-quality defects, defined as any CTRS dimension receiving a score of 3 or lower (\textbf{3.89\%} rejected)~\citep{blackburn2001revised}.
To validate filtering reliability, we randomly sample 100 sessions and ask two licensed experts to independently evaluate the quality-control criteria. The average inter-rater agreement, measured by Cohen’s Kappa~\citep{cohen1960coefficient}, is \(\kappa = 0.46\).

After filtering, DMTCorpus contains 4,317 retained counseling sessions with longitudinal therapeutic progression, multimodal behavioral grounding, tool-augmented intervention events, and cross-session homework continuity.

\begin{table*}[t]
  \centering
  \small
  \setlength{\tabcolsep}{5pt}
  \renewcommand{\arraystretch}{1.15}
  \begin{tabular}{
    l
    S[table-format=1.3]
    S[table-format=1.3]
    S[table-format=1.3]
    S[table-format=1.3]
    S[table-format=1.3]
    S[table-format=1.3]
  }
    \toprule
    \multirow{2}{*}{\textbf{Models}} 
    & \multicolumn{3}{c}{\textbf{General Counseling Skills}} 
    & \multicolumn{3}{c}{\textbf{CBT-specific Skills}} \\
    \cmidrule(r){2-4} \cmidrule(l){5-7}
    & \textbf{Understanding} 
    & \textbf{Interpersonal} 
    & \textbf{Collaboration} 
    & \textbf{Guided Discovery} 
    & \textbf{Focus} 
    & \textbf{Strategy} \\
    \midrule
    \textsc{Camel}     & 4.686$^{*}$  & 4.600        & 4.786$^{*}$  & 4.525$^{*}$  & 4.411$^{*}$  & 4.213$^{*}$ \\
    PsyDTLLM  & 4.736$^{*}$  & 4.650        & 4.767$^{*}$  & 4.497$^{*}$  & 4.400$^{*}$  & 4.202$^{*}$ \\
    \specialrule{0.8pt}{0pt}{0pt}
    CS-LLaVA  & 4.680$^{*}$  & 4.533$^{*}$  & 4.861$^{*}$  & 4.569$^{*}$  & 4.375$^{*}$  & 4.294 \\
    Mirror    & 4.706$^{*}$  & 4.621        & 4.952$^{*}$  & 4.571        & 4.460$^{*}$  & 4.360 \\
    \specialrule{0.8pt}{0pt}{0pt}
    PsychEval & 4.655$^{*}$  & 4.636        & 4.816$^{*}$  & 4.569$^{*}$  & 4.527$^{*}$  & 4.358 \\
    \textbf{Ours} & \textbf{4.936} & \textbf{4.709} & \textbf{5.082} & \textbf{4.781} & \textbf{4.646} & \textbf{4.375} \\
    \bottomrule
  \end{tabular}
    \caption{\label{tab:ctrs_main}
    Session-level CTRS evaluation results. Higher scores indicate better performance.
    The asterisk ($^{*}$) indicates a significant difference compared with Ours
    ($p < 0.05$, paired t-test). 
    }
\end{table*}

\begin{figure*}[t]
    \centering
    \includegraphics[width=0.8\textwidth]{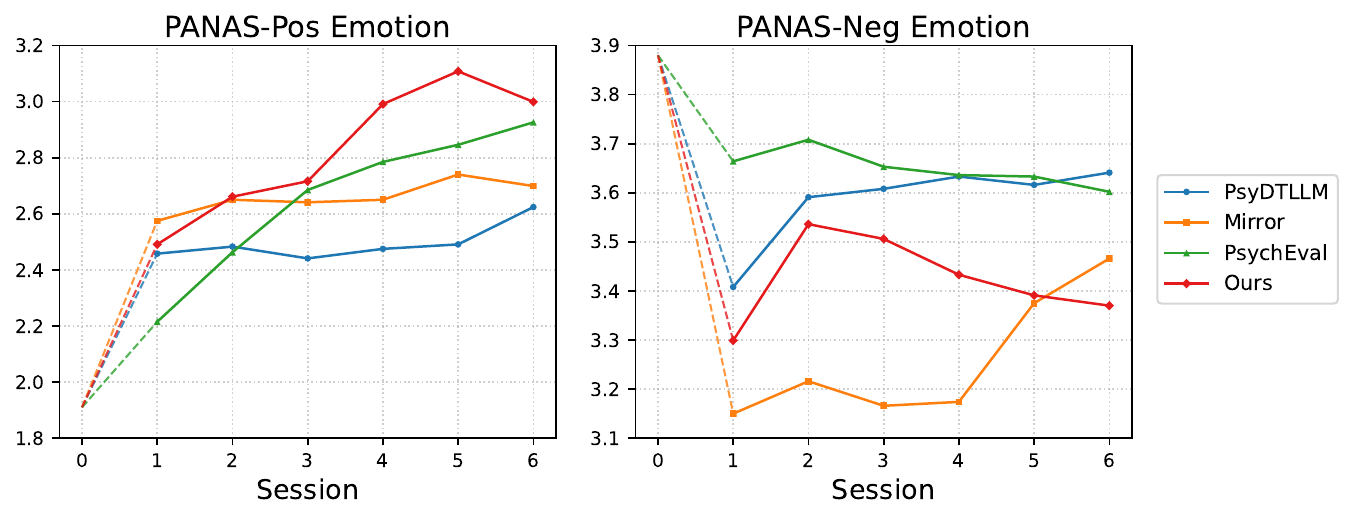}
    \caption{PANAS positive and negative emotion trends across sessions for different methods.}
    \label{fig:panas_curve}
\end{figure*}

\section{Experiments}

\subsection{Experimental Setup}

\paragraph{Dataset and Models.}
We fine-tune the core components of DMT-CBT on the training subset of DMTCorpus, which contains 3,597 retained sessions from 648 case-style conditions. The remaining 720 sessions are used for internal validation. For model comparison, we use the held-out counseling cases described in Appendix~\ref{app:text_resources}, ensuring that evaluation cases are not used during corpus construction.

We adopt Qwen2.5-VL-7B-Instruct~\citep{qwen2024qwen2} as the backbone model for multimodal counseling generation and Qwen2.5-7B-Instruct~\citep{qwen2024qwen2} for therapeutic-state tracking and intervention planning. All components are fine-tuned with LoRA~\citep{hu2022lora}. We further instantiate the client simulation component with GPT-4o-mini for longitudinal multimodal counseling simulation. Detailed hyper-parameters, training schedules, and prompts are provided in Appendix~\ref{app:experimental_setup}.

\paragraph{Baselines.}
We compare DMT-CBT with three categories of baselines:
(1) \textbf{single-session text-only systems}, including \textsc{Camel}~\citep{lee2024cactus} and PsyDTLLM~\citep{xie2025psydt};
(2) \textbf{single-session multimodal systems}, including CS-LLaVA~\citep{kim2025multimodal} and Mirror~\citep{kim2025mirror};
and (3) \textbf{multi-session text-only systems}, including PsychEval~\citep{pan2026psycheval}.

These baselines progressively differ in multimodal capability, longitudinal interaction, and therapeutic-state modeling.

\paragraph{Evaluation Metrics.}
We evaluate DMT-CBT from three perspectives.
For \textbf{session-level therapeutic quality}, we adopt CTRS~\citep{blackburn2001revised} and WAI~\citep{horvath1989development}.
For \textbf{longitudinal therapeutic evolution}, we evaluate cross-session affective trajectories using PANAS~\citep{watson1988development}.
For \textbf{longitudinal therapeutic-state modeling}, we further evaluate therapeutic-stage control, intervention selection, homework recommendation, and memory consistency. GPT-4o-mini and GPT-5-chat are used as automatic evaluators. Detailed evaluation protocols are provided in Appendix~\ref{app:evaluation_method}.

\subsection{Session-Level Therapeutic Evaluation}

We first evaluate whether DMT-CBT improves session-level counseling quality. Tables~\ref{tab:ctrs_main} and~\ref{tab:wai_main} report results on CTRS and WAI, respectively.

On CTRS, DMT-CBT achieves the best performance across all six dimensions, covering both general counseling competence and CBT-specific intervention skills. The improvements are particularly clear on Collaboration, Guided Discovery, Focus, and Strategy, suggesting that structured therapeutic-state tracking improves intervention planning and session organization.

On WAI, DMT-CBT achieves the best performance on Goal and competitive performance on Task and Bond. These results indicate that longitudinal therapeutic-state modeling improves therapeutic alignment and collaborative counseling progression while maintaining strong therapeutic rapport.

Overall, the results demonstrate that DMT-CBT not only improves local response quality but also enhances session-level therapeutic structure and CBT-oriented intervention effectiveness.

\subsection{Longitudinal Therapeutic Evaluation}

We next evaluate whether DMT-CBT effectively models therapeutic progression across sessions. Figure~\ref{fig:panas_curve} shows the longitudinal PANAS trajectories throughout counseling.

Compared with existing systems, DMT-CBT produces a clearer increase in positive affect together with a more stable reduction in negative affect across sessions. PsyDTLLM shows relatively limited positive-affect improvement, while Mirror mainly improves during early sessions before gradually stabilizing. PsychEval exhibits smoother longitudinal trajectories but achieves smaller overall affective gains than DMT-CBT.

These results suggest that DMT-CBT better captures simulated affective trajectories across sessions, extending beyond local session-level response quality. However, since PANAS scores are inferred from synthetic interactions by automatic evaluators, they should be interpreted as affective proxies within controlled simulation settings rather than evidence of real-world emotional improvement.


\begin{table}[t]
  \centering
  \small
  \setlength{\tabcolsep}{7pt}
  \renewcommand{\arraystretch}{1.15}
  \begin{tabular}{
    l
    S[table-format=1.3]
    S[table-format=1.3]
    S[table-format=1.3]
  }
    \toprule
    \textbf{Models} 
    & \textbf{Task ($\uparrow$)} 
    & \textbf{Goal ($\uparrow$)} 
    & \textbf{Bond ($\uparrow$)} \\
    \midrule
    \textsc{Camel}       & 3.327 & 3.427$^{*}$ & 3.648$^{*}$ \\
    PsyDTLLM    & 3.226$^{*}$ & 3.403$^{*}$ & 3.765$^{*}$ \\
    \specialrule{0.8pt}{0pt}{0pt}
    CS-LLaVA    & 3.222$^{*}$ & 3.339$^{*}$ & \textbf{3.831}$^{*}$ \\
    Mirror      & 3.286 & 3.566 & 3.759$^{*}$ \\
    \specialrule{0.8pt}{0pt}{0pt}
    PsychEval   & \textbf{3.340} & 3.635 & 3.627 \\
    \textbf{Ours} 
                & \textbf{3.340} 
                & \textbf{3.798} 
                & 3.711 \\
    \bottomrule
  \end{tabular}
  \caption{\label{tab:wai_main}
    Session-level WAI evaluation results. The asterisk ($^{*}$) indicates a significant difference compared with Ours ($p < 0.05$, paired $t$-test).
  }
\end{table}

\begin{table}[t]
\centering
\small
\setlength{\tabcolsep}{5pt}
\renewcommand{\arraystretch}{1.15}
\begin{tabular}{lcc}
\toprule
\textbf{Module} & \textbf{Metric} & \textbf{Score} \\
\midrule
Stage Modeling & Accuracy & 0.929 \\
State Tracking & Value Sim. & 0.803 \\
Intervention Action Pred. & 3-class Acc. & 0.611 \\
Intervention Action Pred. & Macro-F1 & 0.601 \\
Tool Selection & Top-1 Acc. & 0.964 \\
Homework Rec. & Similarity & 0.833 \\
Hierarchical Memory & Similarity & 0.976 \\
\bottomrule
\end{tabular}
\caption{Evaluation of longitudinal therapeutic-state modeling and intervention capabilities in DMT-CBT.}
\label{tab:system_level}
\end{table}

\begin{table*}[t]
  \centering
  \small
  \setlength{\tabcolsep}{5pt}
  \renewcommand{\arraystretch}{1.15}
  \begin{tabular}{
    l
    S[table-format=1.3]
    S[table-format=1.3]
    S[table-format=1.3]
    S[table-format=1.3]
    S[table-format=1.3]
    S[table-format=1.3]
    S[table-format=2.3]
  }
    \toprule
    \multirow{2}{*}{\textbf{Method}}
    & \multicolumn{3}{c}{\textbf{General Counseling Skills}}
    & \multicolumn{3}{c}{\textbf{CBT-specific Skills}}
    & \multicolumn{1}{c}{\multirow{2}{*}{\textbf{Overall}}} \\
    \cmidrule(r){2-4} \cmidrule(lr){5-7}
    & \multicolumn{1}{c}{\textbf{Understanding}}
    & \multicolumn{1}{c}{\textbf{Interpersonal}}
    & \multicolumn{1}{c}{\textbf{Collaboration}}
    & \multicolumn{1}{c}{\textbf{Guided Discovery}}
    & \multicolumn{1}{c}{\textbf{Focus}}
    & \multicolumn{1}{c}{\textbf{Strategy}}
    & \multicolumn{1}{c}{} \\
    \midrule
    w/o SSM   
      & 4.788$^{*}$ & 4.691 & 4.827$^{*}$ & 4.569 & 4.405$^{*}$ & 4.158$^{*}$ & 27.438$^{*}$ \\
    w/o TAI
      & 4.875 & 4.730 & 5.025 & 4.638 & 4.458$^{*}$ & 4.283 & 28.009 \\
    w/o Vision
      & 4.847 & 4.700 & 5.078 & 4.702 & 4.610 & 4.325 & 28.262 \\
    w/o IDA   
      & 4.812 & {\bfseries 4.781} & 5.056 & 4.693 & 4.593 & 4.356 & 28.291 \\
    w/o IVG  
      & {\bfseries 4.947} & 4.683 & 5.011 & 4.766 & 4.555 & 4.360 & 28.322 \\
    \textbf{Ours (Full)}
      & 4.936 
      & 4.709 
      & {\bfseries 5.082} 
      & {\bfseries 4.781} 
      & {\bfseries 4.646} 
      & {\bfseries 4.375} 
      & {\bfseries 28.529} \\
    \bottomrule
  \end{tabular}
  \caption{\label{tab:ablation_ctrs}
    Ablation results on session-level CTRS evaluation. Vision refers to the combination of IVG and IDA. The asterisk ($^{*}$) indicates a significant difference compared with Ours ($p < 0.05$, paired t-test).
  }
\end{table*}

\begin{figure*}[t]
    \centering
    \includegraphics[width=\textwidth]{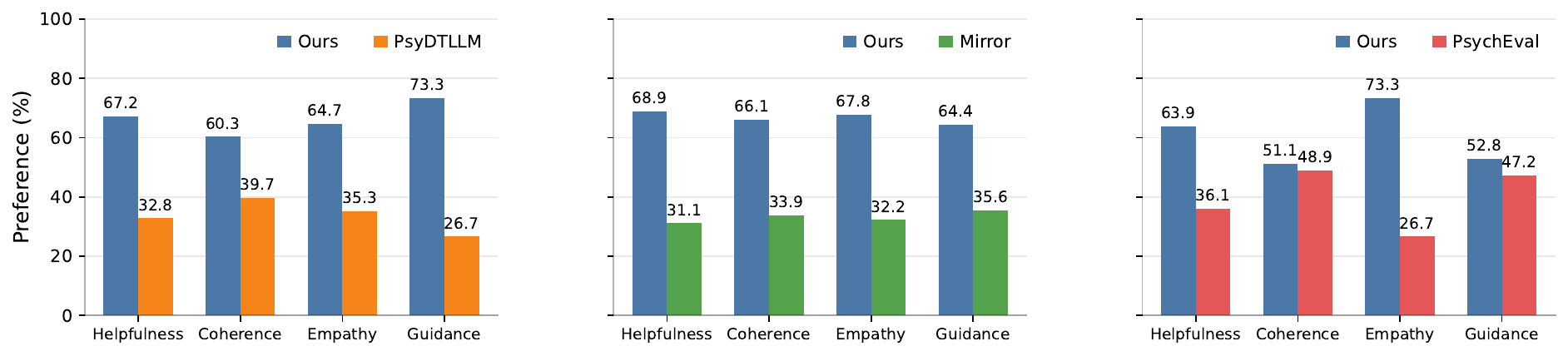}
    \caption{Expert evaluation results shown as pairwise preference percentages. All results demonstrate statistically significant differences with p < 0.05, except for the Coherence and Guidance between Ours and PsychEval.}
    \label{fig:human_eval}
\end{figure*}

\subsection{Longitudinal State Analysis}

We further evaluate whether DMT-CBT maintains structured therapeutic states throughout longitudinal counseling. Table~\ref{tab:system_level} reports the reliability of the framework's therapeutic-state modeling and intervention components.

The therapeutic-stage modeling component achieves high stage-transition accuracy and state-value similarity, indicating that the framework can track CBT progression and structured counseling states. The intervention planning component achieves high tool-selection accuracy, while homework recommendation remains consistent with reference assignments in the simulation protocol.

The hierarchical memory mechanism achieves higher memory consistency than direct post-hoc extraction from full-session dialogue history. To further assess whether this advantage transfers to therapist responses, we conduct a response-level cross-session memory analysis in Appendix~\ref{app:memory_eval}. This analysis compares three memory implementations, namely w/o memory, post-hoc extraction, and DMT-CBT memory, under the same response-generation setting. Results show that DMT-CBT memory achieves stronger State Recall, Homework Continuity, and Response Grounding, while reducing response-level errors such as contradictions, hallucinated prior events, and important omissions.

These findings suggest that explicitly maintaining structured counseling states supports more grounded cross-session response generation than relying only on raw dialogue context or post-hoc summarization.

\subsection{Ablation Study}
We examine the contribution of four key components: Structured State Modeling (SSM), Tool-Augmented Intervention (TAI), Image-based Visual Grounding (IVG), and Image Description Augmentation (IDA). Table~\ref{tab:ablation_ctrs} shows that removing any component degrades overall counseling quality. Removing SSM causes the largest overall performance drop, suggesting that explicit therapeutic-state modeling is critical for longitudinal counseling modeling. Removing TAI mainly reduces CBT-specific intervention quality, while removing Vision causes a larger degradation than removing IVG or IDA alone, suggesting that image-based visual grounding and image-description augmentation provide complementary support for modeling nonverbal cues. These results demonstrate that DMT-CBT benefits jointly from structured state propagation, multimodal grounding, and tool-augmented intervention.

To further analyze multimodal behavioral grounding, we conduct a targeted nonverbal incongruence diagnostic in Appendix~\ref{app:incongruence}. The diagnostic evaluates whether therapist responses can recognize and cautiously handle mismatches between client verbal content and nonverbal cues. DMT-CBT achieves the best results on IAR, CCR, NCU, CBT Utility, and nCRSS, while obtaining the lowest OIR under both judges. These results indicate stronger sensitivity to verbal--nonverbal mismatch, better use of nonverbal cues for CBT-oriented exploration, and lower unsupported over-interpretation. The consistent ranking across the two judge models further suggests robustness to evaluator choice.

\subsection{Expert Evaluation}We further conduct expert evaluation with two licensed professionals on 30 counseling cases covering 180 sessions in total. Each session is independently rated by both experts. The experts compare DMT-CBT with PsyDTLLM, Mirror, and PsychEval on helpfulness, coherence, empathy, and guidance~\citep{lee2024cactus}. As shown in Figure~\ref{fig:human_eval}, experts consistently prefer DMT-CBT over all baselines on most dimensions, with clearer advantages on helpfulness, empathy, and guidance and a smaller margin on coherence, especially against PsychEval. The largest gains are observed on helpfulness and guidance, indicating that longitudinal therapeutic-state modeling improves sustained counseling progression and intervention relevance across sessions. The inter-rater agreement is moderate (\(\kappa = 0.51\)).

%

\section{Conclusion}

In this paper, we propose DMT-CBT, a framework for longitudinal therapeutic state modeling in CBT counseling. DMT-CBT models psychotherapy as a dynamic and partially observable process involving multimodal client understanding, structured intervention planning, and cross-session therapeutic-state propagation. We further construct DMTCorpus, a multi-session multimodal CBT counseling dataset with image-grounded behavioral cues, tool-augmented intervention events, and longitudinal homework continuity.
Experimental results demonstrate that DMT-CBT provides empirical evidence within controlled simulation settings, and preserves therapeutic states more faithfully than post-hoc memory extraction approaches. We hope this work can provide a foundation for future research on longitudinal psychotherapy modeling and clinically grounded LLM-based mental health systems.

\section*{Limitations}

This work has several limitations. First, DMT-CBT is evaluated mainly in simulated counseling settings rather than real clinical deployment, so it may not fully capture the complexity and unpredictability of real therapeutic interactions. Second, DMTCorpus is generated through LLM-based simulation and several evaluations rely on LLM judges, which may introduce evaluation circularity and model-family bias. Therefore, our results should be interpreted as relative evidence within controlled synthetic settings, rather than evidence of real-world therapeutic efficacy. Third, the current multimodal signals are primarily based on still facial-expression images, leaving richer temporal and nonverbal cues such as gesture, posture, and vocal prosody underexplored. Finally, DMTCorpus is grounded mainly in a Chinese linguistic and cultural context, which may limit its generalizability across populations, counseling cultures, and therapeutic settings. Future work will extend the framework to richer video-based multimodal signals, more diverse cultural contexts, and more clinically grounded evaluation protocols.

\section*{Ethical Considerations}

\paragraph{Data Privacy.} DMTCorpus is constructed through simulated counseling interactions rather than direct collection of real client--therapist conversations. The textual counseling sessions, homework interactions, and therapist-side memory states are generated by instantiating the proposed multi-agent framework, which reduces the risk of exposing personally identifiable information. For visual resources, the client image library is derived from DISFA, a publicly available facial-expression dataset, and is used only to retrieve nonverbal cues for simulation rather than to identify real individuals.

\paragraph{Clinical Safety.} DMT-CBT is designed as a research framework for studying multimodal, multi-session CBT counseling simulation, not as a standalone clinical system. Although the framework incorporates CBT principles, structured tools, and memory tracking, its outputs may still contain errors or inappropriate suggestions due to the uncertainty of LLM generation. Moreover, users or downstream developers may overtrust it or misuse it as quasi-therapy. Therefore, the system should be used only as a supportive tool under the supervision of qualified mental health professionals.  It is not intended for crisis intervention, diagnosis, or the treatment of high-risk clinical cases. The simulated affective trajectories and counseling-quality scores reported in this paper should not be interpreted as clinically meaningful therapeutic progress.

\paragraph{Bias and Generalization.} We acknowledge potential demographic, cultural, and linguistic biases in both the textual and visual resources. DMTCorpus is primarily grounded in a Chinese language and cultural context, which may limit its generalizability to other populations or counseling practices. Future work should expand the dataset to more diverse linguistic, cultural, and demographic settings and conduct more systematic bias evaluation.

\bibliography{custom}

\appendix

\section{Cognitive Behavioral Therapy}
\label{app:cbt_background}

\subsection{Overview}

Cognitive Behavioral Therapy (CBT) is a structured, goal-oriented, and evidence-based psychotherapy approach. It assumes that emotional and behavioral difficulties are closely related to how individuals interpret situations. CBT therefore aims to help clients identify maladaptive thoughts and beliefs, evaluate their validity, and develop more adaptive coping strategies. Compared with open-ended supportive counseling, CBT typically follows a structured treatment process across multiple sessions.

\subsection{Core Components}

\paragraph{Cognitive Conceptualization Diagram.}
The Cognitive Conceptualization Diagram (CCD) is a structured formulation tool for representing a client's cognitive, emotional, and behavioral patterns. A typical CCD consists of eight components: relevant history, core beliefs, intermediate beliefs, compensatory strategies, situations, automatic thoughts, emotions, and behaviors. These components provide a concise representation of how prior experiences and cognitive schemas shape current responses. An example is shown in Table~\ref{tab:ccd_example}.

\paragraph{Therapeutic Tools.}
CBT commonly employs structured tools such as thought records, activity schedules, behavioral experiments, belief worksheets, and psychoeducational diagrams. These tools help clients externalize internal experiences, examine evidence, and practice therapeutic skills in concrete forms.

\paragraph{Homework.}
Homework is a key mechanism for extending therapeutic work beyond the session. Typical assignments include monitoring automatic thoughts, recording emotions, testing beliefs, scheduling activities, and practicing coping strategies. Homework is usually reviewed in subsequent sessions, thereby supporting continuity across treatment.

\subsection{Session Structure}

CBT is typically organized as a multi-session process. The first session often focuses on assessment, including background information, presenting concerns, current difficulties, and initial treatment goals. Later sessions generally progress from psychoeducation and the cognitive model to automatic-thought intervention, intermediate beliefs, core beliefs, and relapse prevention.

At the session level, CBT usually follows three broad phases: assessment and agenda setting, intervention and skill practice, and summary with homework planning. Although the exact structure varies across clients and treatment goals, CBT emphasizes collaboration, goal orientation, structured intervention, between-session practice, and longitudinal continuity.

\section{DMTCorpus Construction}
\label{app:dmtcorpus}

\subsection{Text Resources}
\label{app:text_resources}

\paragraph{CBT Treatment Structure.}
We define a generic multi-session CBT treatment structure to represent the clinical progression of therapy. The six-session playbook follows a progressive CBT structure:
\begin{itemize}
    \item \textbf{Session 1: Intake and treatment planning.} The therapist collects recent status, daily routine, history, initial goals, treatment plan, and treatment expectations.
    \item \textbf{Session 2: Problem specification and cognitive education.} The therapist reviews recent emotions, identifies concrete problems, transforms problems into goals, introduces behavioral planning, and elicits automatic thoughts.
    \item \textbf{Session 3: Automatic thought.} The therapist reviews homework, identifies negative and positive events, evaluates automatic-thought belief strength, searches for counter-evidence, and constructs alternative thoughts.
    \item \textbf{Session 4: Intermediate beliefs.} The therapist uses downward-arrow questioning, evaluates the advantages and disadvantages of intermediate beliefs, and constructs alternative intermediate beliefs.
    \item \textbf{Session 5: Core beliefs.} The therapist identifies core beliefs, links them to early experiences, explores compensatory strategies, and constructs alternative core beliefs.
    \item \textbf{Session 6: Consolidation and relapse prevention.} The therapist reviews the cognitive model, reinforces progress, creates a self-therapy plan, and prepares relapse-prevention strategies.
\end{itemize}

Within each session, the structure follows three broad phases: assessment, intervention, and summary.

\paragraph{Homework Repository.}
We build a structured homework repository to support between-session CBT practice. Each homework item is associated with metadata such as category, difficulty, and applicable stage. The category label follows a six-class taxonomy: Movement, Work and Education, Spare Time, Daily Living, Practical Tasks, and Social Activities. Table~\ref{tab:homework_category_stats} summarizes the category distribution of the current homework repository. 

Homework is generated with a two-stage pipeline. The system first retrieves top-$k$ candidate activities from homework repository using the current state and recent client reply as the query. Retrieval is based on embedding similarity, with a token-overlap fallback.

\begin{table}[t]
\centering
\small
\begin{tabular}{lc}
\toprule
\textbf{Homework Category} & \textbf{\# Items} \\
\midrule
Movement & 77 \\
Work and Education & 28 \\
Spare Time & 128 \\
Daily Living & 29 \\
Practical Tasks & 44 \\
Social Activities & 77 \\
\midrule
Total & 383 \\
\bottomrule
\end{tabular}
\caption{Category distribution of the homework repository.}
\label{tab:homework_category_stats}
\end{table}

\subsection{Multimodal Resources}
\label{app:mm_resources}

\paragraph{Image Library.}
The client image library is constructed from a facial-expression video dataset with frame-level AU annotations and seven-way emotion labels. Since our goal is to provide controlled nonverbal cues rather than model continuous facial dynamics, we convert the original video sequences into a static image library through interval-based frame sampling. This strategy reduces near-duplicate frames while preserving diverse facial states. Each retained frame is associated with structured metadata, including subject identity, AU-based facial information, an emotion label, and textual descriptors of the observed expression or action. These descriptors make the visual cues interpretable and retrieval-friendly, allowing them to be used either as image inputs or textual grounding.

\paragraph{Tool Library.}\label{app:mm_tool}
The visual CBT tool library contains structured therapeutic materials such as worksheets, diagrams, psychoeducational aids, and intervention templates. Each tool is associated with metadata describing its applicable stages, triggering conditions, therapist instructions, usage examples, textual content, and visual material. 

Tool retrieval is further constrained by a \textbf{stage-step whitelist strategy}. Instead of searching over the entire library at every turn, the system first checks whether the current session step belongs to a predefined whitelist of tool-eligible CBT stages, such as cognitive-model introduction, problem concretization, automatic-thought work, homework assignment, homework review, and relapse prevention. Only tools permitted for the current stage are considered as candidates. This design reduces irrelevant tool activation and keeps tool use aligned with the intended therapeutic structure.

After whitelist filtering, the remaining candidates are ranked by semantic relevance using embedding-based retrieval. The ranking combines stage information and local dialogue context, allowing the selected tool to be both \emph{stage-appropriate} and \emph{context-sensitive}.

\subsection{Details of DMT-CBT}
\paragraph{Multimodal Replacement Strategy.}\label{app:replacement}
We introduce a controlled \emph{multimodal incongruence} mechanism to make simulated client behavior more realistic. Specifically, the client's nonverbal signal is occasionally replaced with an emotion--action state that does not fully align with the verbal content.

The replacement is phase-aware. Each session is divided into early, middle, and late phases, with incongruence trigger probabilities of 0.10, 0.30, and 0.20, respectively. No incongruence is injected at the first turn. When triggered, the incongruence strength is sampled from three levels: mild, medium, and strong.

Once activated, the original emotion label is replaced by a new emotion sampled from the seven-category emotion set. The sampling is relation-aware: emotions that create stronger incongruence with the original label receive the highest probability (0.7), mildly incongruent emotions receive a moderate probability (0.2), and other emotions receive a lower probability (0.1). A corresponding facial-action description is then sampled from the action pool associated with the new emotion. To reduce repetition, we use a deduplication window of 6 for action templates.

Unlike MIRROR-style stage-direction generation, our strategy explicitly controls and annotates verbal--nonverbal incongruence. This makes multimodal inconsistency controllable, interpretable, and analyzable rather than incidental.

This strategy introduces interpretable, stage-aware nonverbal deviations rather than random noise, making therapist-side multimodal grounding more challenging and closer to realistic counseling interactions. The relation-aware emotion mapping is provided in Table~\ref{tab:emotion_relation_mapping}.

\paragraph{Client Image Matching.}\label{app:match}
Each dialogue client is first mapped to a stable subject in the facial-expression library to maintain cross-session visual consistency. Subject assignment is constrained by gender and age metadata, preferring same-gender matches and filtering candidates within an age window before deterministic selection. 

At each turn, image retrieval is restricted to the subset matching the client’s current emotion. The query is built from the client’s current emotion and action description, and candidate images are ranked by embedding similarity using metadata-derived text representations. One image is then deterministically selected from the top-ranked set. If the locked subject has no image under the current emotion, retrieval falls back to the full image pool for that emotion.

\paragraph{Client Styles.}\label{app:styles}
To diversify client behaviors, we assign each simulated client one of six interpersonal styles. These styles are designed to reflect different counseling-relevant relational tendencies.

\begin{itemize}
    \item \textbf{Feeling misunderstood}: The client perceives the therapist as procedural or insufficiently empathic, resulting in psychological distance and a sense of alienation.
    \item \textbf{Feeling controlled}: The client becomes dissatisfied with the therapist's suggestions or arrangements, leading to oppositional emotions and behaviors.
    \item \textbf{Defensive or suspicious}: The client uses vague or evasive language to conceal information and prevent the therapist from fully understanding the situation.
    \item \textbf{Controlling}: The client attempts to dominate the counseling relationship and expects the therapist to act according to their preferences.
    \item \textbf{Passive compliance}: The client shows limited autonomy and frequently seeks advice, reassurance, or explicit guidance from the therapist.
    \item \textbf{Externalizing responsibility}: The client tends to shift responsibility away from themselves and is especially concerned that failures may be attributed to their own actions.
\end{itemize}

\subsection{Functional Component Details}

Table~\ref{tab:agent_details} summarizes the role-specialized functional components used to instantiate DMT-CBT. These components implement the longitudinal therapeutic-state modeling framework described in Section~\ref{method}, rather than defining a separate multi-agent contribution. This decomposition makes the simulation and generation process more interpretable while keeping the central design focused on therapeutic-state tracking, multimodal grounding, and longitudinal memory propagation.

\paragraph{Client Agent}
Figure~\ref{fig:client_prompt} shows the unified prompt template for the Client Agent~\citep{wang-etal-2024-patient}. The first-session prompt is profile-based, whereas later-session prompts are conditioned on session-specific cognitive conceptualization.

\paragraph{Therapist Agent} 
Figure~\ref{fig:therapist_prompt} shows the prompt template for the Therapist Agent.

\paragraph{Strategy Agent}
Figure~\ref{fig:strategy_prompt} shows the prompt template for the Strategy Agent.

\paragraph{Tool Agent}
Figure~\ref{fig:tool_prompt} shows the unified prompt template for the Tool Usage. Candidate tools are retrieved in advance from the tool library. The Tool Agent then selects the prompt according to the current tool state: when no tool is active, \texttt{tool\_prompt\_trigger} determines whether to trigger a tool and which candidate tool to select; when a tool is active, \texttt{tool\_prompt\_over} determines whether to terminate the current tool. This design separates candidate retrieval from state-dependent tool scheduling. Figure~\ref{fig:homework_prompt} shows the prompt template for the Homework Module.

\section{Evaluation Method}
\label{app:evaluation_method}

We evaluate DMT-CBT from three perspectives: session-level clinical quality, case-level affective change, and expert human preference. Automatic evaluations are conducted with GPT-4o-mini. PANAS evaluation is conducted with GPT-5-Chat. 

\subsection{Cognitive Therapy Rating Scale (CTRS)}
\label{app:ctrs}

The Cognitive Therapy Rating Scale (CTRS) is used to assess clinical competence in CBT counseling. Following prior work, we adopt a streamlined six-dimension subset that can be reliably observed from generated dialogues. The selected dimensions include three general counseling skills: \textit{Understanding}, \textit{Interpersonal Effectiveness}, and \textit{Collaboration}; and three CBT-specific skills: \textit{Guided Discovery}, \textit{Focus}, and \textit{Strategy}.
Each dimension is scored on a 6-point scale. The prompt used to guide CTRS scoring is shown in Figure~\ref{fig:CTRS}.

\subsection{Working Alliance Inventory (WAI)}
\label{app:wai}
The Working Alliance Inventory (WAI) is used to evaluate the therapeutic alliance between the counselor and the client. Following prior dialogue evaluation settings, we adopt the 12-item WAI and score each item on a 5-point scale. The items are grouped into three dimensions: \textit{Goal}, which measures agreement on counseling objectives; \textit{Task}, which measures agreement on therapeutic activities; and \textit{Bond}, which measures the relational connection between the counselor and the client. The prompt used to guide WAI scoring by is shown in Figure~\ref{fig:WAI}.

\subsection{Positive and Negative Affect Schedule (PANAS)}
\label{app:panas}

The Positive and Negative Affect Schedule (PANAS) is used to assess case-level affective change across sessions. PANAS contains two 10-item subscales: positive affect and negative affect. The positive affect subscale includes emotions such as interested, excited, strong, enthusiastic, proud, alert, inspired, determined, attentive, and active. The negative affect subscale includes emotions such as distressed, upset, guilty, scared, hostile, irritable, ashamed, nervous, jittery, and afraid. Each item is rated on a 5-point scale, where higher scores indicate stronger affective intensity.

We adapt PANAS as a longitudinal emotional assessment tool. For each case, we first conduct a pre-counseling assessment at Session~0 based on the client profile, which serves as the affective baseline. GPT-5-Chat then infers the client's affective state after each counseling session from the dialogue content and available client-state information. This setup allows us to track changes in positive and negative affect from the initial profile-based baseline through the subsequent multi-session counseling process. The structured prompt guiding this scoring process is shown in Figure~\ref{fig:panas}.

\subsection{Expert Evaluation}
\label{app:expert_eval}

To further assess session quality, two licensed professionals with formal clinical qualifications conduct expert evaluation. The evaluation covers four counseling-related dimensions: \textit{Helpfulness}, \textit{Empathy}, \textit{Logical Coherence}, and \textit{Guidance}. The detailed scoring criteria for \textit{Empathy}, \textit{Logical Coherence}, and \textit{Guidance} follow~\citet{xiao-etal-2024-healme}; the \textit{Helpfulness} criterion is adapted to emphasize therapeutic progress within the DMT-CBT framework.

The experts compare DMT-CBT with representative baselines in a pairwise manner. For each comparison, they judge which session is better on each dimension. Inter-rater agreement is measured using Kappa. Annotators were recruited via a professional crowdsourcing platform and were compensated at a rate of 6 RMB per data item. The adapted \textit{Helpfulness} criterion is shown in Figure~\ref{fig:humanmanual}.

\section{Experimental Setup}
\label{app:experimental_setup}

\subsection{Training}
\label{app:training}

We fine-tune all locally deployed agents and baselines with LoRA on frozen backbones. For text-based agents and baselines, we use Qwen2.5-7B-Instruct as the base model. For the multimodal Therapist Agent, we use Qwen2.5-VL-7B-Instruct. The LoRA target modules are \texttt{q\_proj}, \texttt{k\_proj}, \texttt{v\_proj}, \texttt{o\_proj}, \texttt{up\_proj}, \texttt{down\_proj}, and \texttt{gate\_proj}, with rank $r=16$, scaling factor $\alpha=32$, and dropout rate $0.05$.

For DMT-CBT, we train separate modules for therapist response generation, strategy control, tool selection, and homework recommendation. Text-based agents share the same optimization settings, while the multimodal Therapist Agent is trained on image-text structured examples converted into Qwen-VL format. Images are optionally resized to at most $384 \times 288$ for efficient training.

The training configuration for  agents is shown below:

\begin{lstlisting}[language=Python, basicstyle=\ttfamily\small, frame=single, rulecolor=\color{black}, backgroundcolor=\color{gray!10}]
max_length = 4096
learning_rate = 2e-5
per_device_train_batch_size = 2
gradient_accumulation_steps = 8
gradient_checkpointing = True
num_train_epochs = 2
weight_decay = 0.01
lr_scheduler_type = "cosine"
seed = 42
fp16 = True
save_strategy = "epoch"
save_total_limit = 2
\end{lstlisting}

\subsection{Inference}
\label{app:inference}

During inference, locally deployed agents use autoregressive generation with role-specific decoding settings. The Therapist Agent uses \texttt{max\_new\_tokens=512}, \texttt{temperature=0.7}, \texttt{top\_p=0.9}, and \texttt{do\_sample=True} to balance naturalness and diversity. In contrast, the Strategy Agent and Tool Agent use the same token budget but disable sampling with \texttt{do\_sample=False}, since their outputs require stable structured prediction. The Homework Agent uses a smaller generation budget of \texttt{max\_new\_tokens=256}.

System-level inference is further constrained by the CBT treatment structure, tool-stage compatibility, cross-session memory, and the multimodal replacement strategy. Therefore, generation is not treated as unconstrained dialogue continuation, but as structured multi-agent counseling simulation.

\section{Cross-session Memory Effectiveness}
\label{app:memory_eval}
To evaluate longitudinal continuity, we conduct a response-level cross-session memory analysis on held-out cases. We use 20 held-out clients, each instantiated with six interaction styles and six sessions. For each client-style trajectory, we construct five adjacent transitions from Session 1$\rightarrow$2 to Session 5$\rightarrow$6, and select one therapist response in session $k+1$ that requires prior-session information, resulting in 600 samples.

For each transition, the structured \texttt{state\_after} field after session $k$ is used as the reference prior state. We compare three response-generation settings: \textit{w/o memory}, where the therapist receives no prior-session information; \textit{post-hoc extraction}, where GPT-4.1-mini first extracts memory from the full transcript of session $k$ and the therapist then generates the response; and \textit{DMT-CBT memory}, where the response is generated using memory incrementally updated during counseling and propagated to the next session.

We use GPT-4o-mini and Qwen3-Max as judges. Given the reference prior state, current-session client context, and generated therapist response, each judge scores three dimensions on a 1--5 scale: \textbf{State Recall} for preserving prior CBT states, \textbf{Homework Continuity} for following up previous homework or action plans, and \textbf{Response Grounding} for using prior-session progress rather than generic context. We also report \textbf{Response Error}, defined as the proportion of responses containing at least one contradiction, hallucinated prior event, or clinically important omission.

As shown in Table~\ref{tab:memory_eval}, \textit{w/o memory} performs poorly under both judges because the therapist lacks prior-session information. Post-hoc extraction improves continuity but still produces more response-level errors. In contrast, DMT-CBT memory achieves the best scores and the lowest Response Error under both judges, showing that incremental memory tracking better grounds therapist responses in prior-session progress and remains robust to evaluator choice.

\section{Nonverbal Incongruence Analysis} \label{app:incongruence}

To examine whether models can handle mismatches between verbal content and nonverbal cues, we conduct a diagnostic analysis on language--nonverbal incongruence cases. For each method, we randomly sample 180 incongruent instances from the held-out evaluation setting. Each instance contains the dialogue context, the client's verbal response, the nonverbal cue description, and the therapist response.

We evaluate five dimensions with GPT-4o-mini and Qwen3-Max: Incongruence Awareness Rate (IAR), Cautious Clarification Rate (CCR), Nonverbal Cue Utilization (NCU), CBT Utility, and Over-Interpretation Rate (OIR). IAR, CCR, and OIR are binary metrics. IAR measures whether the response recognizes the verbal--nonverbal mismatch; CCR measures cautious and open handling of the cue; OIR measures unsupported over-interpretation of hidden client states. NCU and CBT Utility are rated on a 1--5 scale. We also report normalized Cue-Response Semantic Similarity (nCRSS), computed with BGE-large-zh-v1.5. The detailed prompt is shown in Figure~\ref{fig:prompt_nonverbal_incongruence}.

As shown in Table~\ref{tab:nonverbal_incongruence}, DMT-CBT achieves the best overall results under both judge models. With GPT-4o-mini, it obtains the highest IAR, CCR, NCU, CBT Utility, and nCRSS, while also achieving the lowest OIR. The same pattern holds with Qwen3-Max, where DMT-CBT shows especially clear gains on IAR and CCR. These results indicate stronger sensitivity to verbal--nonverbal mismatch and better cautious use of nonverbal cues for CBT-oriented exploration.

This trend is consistent with the model designs. CS-LLaVA can process visual information but is not explicitly optimized for therapeutic incongruence reasoning. Mirror uses emotion and action cues as auxiliary affective context, whereas DMT-CBT explicitly introduces and annotates controlled incongruence during multimodal grounding. The consistently low OIR further suggests that DMT-CBT improves cue awareness without increasing unsupported interpretation. The consistent ranking across GPT-4o-mini and Qwen3-Max also indicates robustness to evaluator choice.


\begin{table*}[t]
\centering
\small
\setlength{\tabcolsep}{5pt}
\renewcommand{\arraystretch}{1.15}
\begin{tabular}{llcccc}
\toprule
\textbf{Judge Model} & \textbf{Method}
& \textbf{State Recall $\uparrow$}
& \textbf{Homework $\uparrow$}
& \textbf{Grounding $\uparrow$}
& \textbf{Resp. Error $\downarrow$} \\
\midrule
\multirow{3}{*}{GPT-4o-mini}
& w/o memory          & 2.28 & 2.42 & 2.35 & 1.00 \\
& Post-hoc extraction & 3.78 & 3.76 & 4.06 & 0.54 \\
& DMT-CBT memory      & \textbf{4.48} & \textbf{4.00} & \textbf{4.24} & \textbf{0.38} \\
\midrule
\multirow{3}{*}{Qwen3-Max}
& w/o memory          & 1.00 & 1.88 & 1.44 & 1.00 \\
& Post-hoc extraction & 3.52 & 4.00 & 4.28 & 0.42 \\
& DMT-CBT memory      & \textbf{4.00} & \textbf{4.40} & \textbf{4.36} & \textbf{0.34} \\
\bottomrule
\end{tabular}
\caption{
Response-level cross-session memory effectiveness analysis under different judge models. State Recall, Homework Continuity, and Response Grounding are rated on a 1--5 scale. Resp. Error denotes the proportion of generated responses with at least one contradiction, hallucinated prior event, or clinically important omission.
}
\label{tab:memory_eval}
\end{table*}


\begin{table*}[t]
\centering
\small
\setlength{\tabcolsep}{4pt}
\renewcommand{\arraystretch}{1.15}
\begin{tabular}{llcccccc}
\toprule
\textbf{Judge Model} & \textbf{Method} 
& \textbf{IAR} $\uparrow$ 
& \textbf{CCR} $\uparrow$ 
& \textbf{NCU} $\uparrow$ 
& \textbf{CBT Utility} $\uparrow$ 
& \textbf{nCRSS} $\uparrow$ 
& \textbf{OIR} $\downarrow$ \\
\midrule
\multirow{3}{*}{GPT-4o-mini}
& CS-LLaVA  & 0.067 & 0.067 & 2.000 & 2.433 & 0.398 & 0.100 \\
& Mirror    & 0.033 & 0.033 & 1.667 & 2.467 & 0.448 & 0.067 \\
& DMT-CBT   & \textbf{0.233} & \textbf{0.233} & \textbf{2.133} & \textbf{2.700} & \textbf{0.540} & \textbf{0.033} \\
\midrule
\multirow{3}{*}{Qwen3-Max}
& CS-LLaVA  & 0.033 & 0.167 & 1.400 & 2.433 & 0.398 & 0.133 \\
& Mirror    & 0.000 & 0.433 & 1.500 & 2.933 & 0.448 & 0.067 \\
& DMT-CBT   & \textbf{0.300} & \textbf{0.600} & \textbf{2.133} & \textbf{3.167} & \textbf{0.540} & \textbf{0.033} \\
\bottomrule
\end{tabular}
\caption{Diagnostic evaluation of therapist responses to language--nonverbal incongruence. IAR, CCR, and OIR are binary rates; NCU and CBT Utility are rated on a 1--5 scale; nCRSS measures semantic alignment between nonverbal cue descriptions and therapist responses.}
\label{tab:nonverbal_incongruence}
\end{table*}

\section{Error Analysis}
\subsection{Failure Modes}
\label{app:failure_modes}

We observe several potential failure modes:
\begin{itemize}
    \item \textbf{Tool-action ambiguity:} the selected tool ID can be correct, but deciding whether to trigger, maintain, or terminate the tool remains difficult.
    \item \textbf{Repetitive empathic openings:} therapist responses may overuse similar supportive phrases.
    \item \textbf{Negative-affect regulation:} PANAS results suggest that increasing positive affect is easier than consistently reducing negative affect.
\end{itemize}

\subsection{Failure Cases}
\label{sec:failure_cases}

\begin{figure*}[t]
    \centering
    \includegraphics[width=\textwidth]{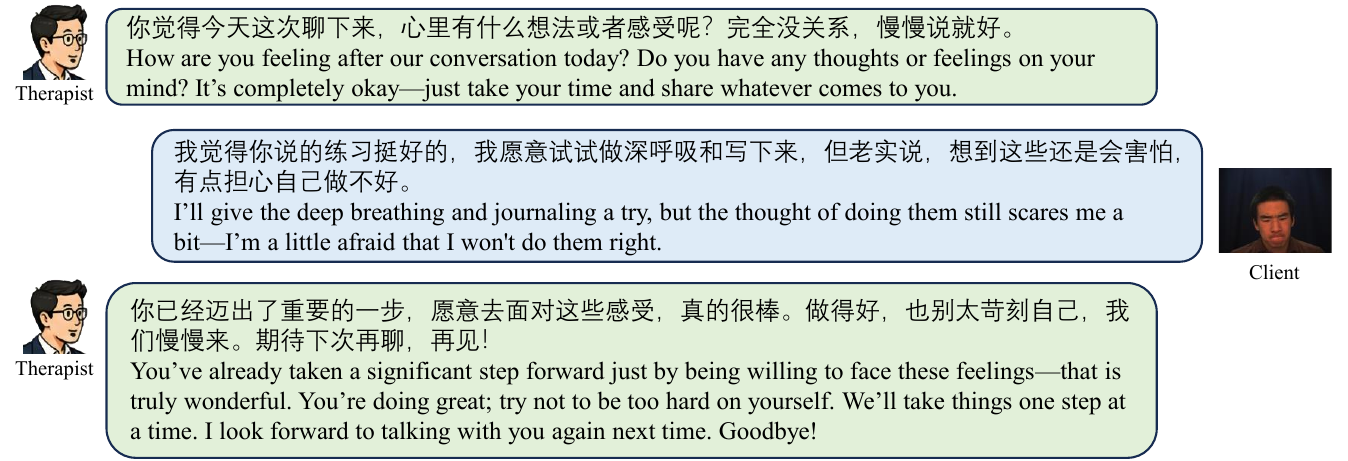}
    \caption{A low-CTRS example during session closure. }
    \label{fig:failure_case_ctrs}
\end{figure*}

Figure~\ref{fig:failure_case_ctrs} shows a representative low-CTRS case that occurs during the session summary stage. At this point, the therapist has already assigned homework and begins to close the session. However, the client expresses dissatisfaction and concern about the assigned exercises, stating that breathing practice and journaling still feel difficult and that they may not complete them correctly. Instead of returning to the homework discussion, clarifying the barrier, or collaboratively modifying the assignment, the therapist mainly provides reassurance and proceeds with a general summary.

This case reflects a limitation in flexible stage control. Although the response remains supportive, the therapist follows the planned summary stage rigidly and fails to respond to a clinically relevant objection raised by the client. Future work should improve the model's ability to detect late-stage client resistance and dynamically return to the appropriate therapeutic step before closing the session.

\section{Case Study}
\label{app:case_study}

\subsection{Case Overview}
\label{app:case_overview}
To illustrate how therapeutic states are progressively constructed across sessions, we present a case-level example based on the structured state updates extracted by the Strategy Agent. Table~\ref{tab:case_state_updates} summarizes the key fields updated in each session. The extracted states follow the generic CBT treatment progression: the first session focuses on intake assessment and treatment planning; the middle sessions gradually move from recent problems and automatic thoughts to intermediate and core beliefs; and the final session consolidates the client's cognitive model and prepares relapse-prevention strategies. These structured updates form a reusable counseling record that supports longitudinal continuity across sessions.

\subsection{Multimodal Replacement}
\label{app:case_multimodal_replacement}

\begin{figure*}[t]
    \centering
    \includegraphics[width=\linewidth]{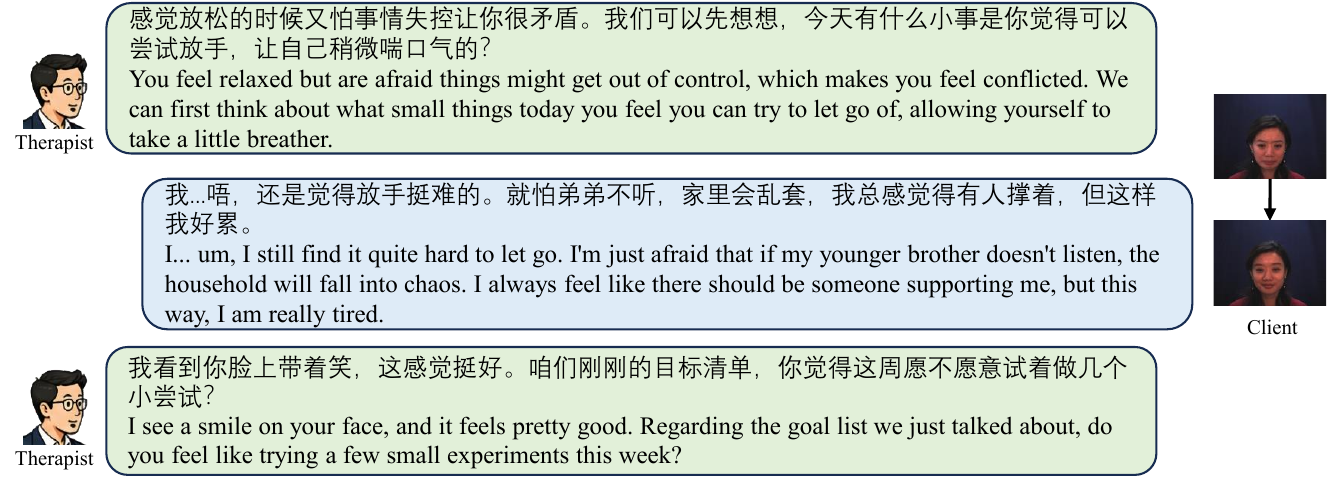}
    \caption{
    Example of multimodal replacement. The client's verbal content conveys sadness and fatigue, while the nonverbal cue is replaced with a smiling expression, producing a controlled sadness-to-happiness incongruence.
    }
    \label{fig:case_multimodal_replacement}
\end{figure*}

Figure~\ref{fig:case_multimodal_replacement} illustrates an example of multimodal replacement. In this case, the client's verbal response conveys a negative emotional state: she reports that letting go is difficult, worries that the household may fall into chaos, and states that she feels tired. However, the associated nonverbal cue is replaced from a sad facial expression with a smiling expression, creating a controlled sadness-to-happiness incongruence.

This example shows how multimodal replacement introduces nonverbal signals that are not fully aligned with the literal verbal content. Instead of treating the client state as directly observable from text alone, the therapist must reason over potentially conflicting cues. In the dialogue, the therapist explicitly responds to the smile while continuing the goal-related discussion, demonstrating how the injected cue can influence the subsequent therapeutic interaction.

\subsection{Tool Usage}
\label{app:case_tool_usage}

\begin{figure*}[t]
    \centering
    \includegraphics[width=\linewidth]{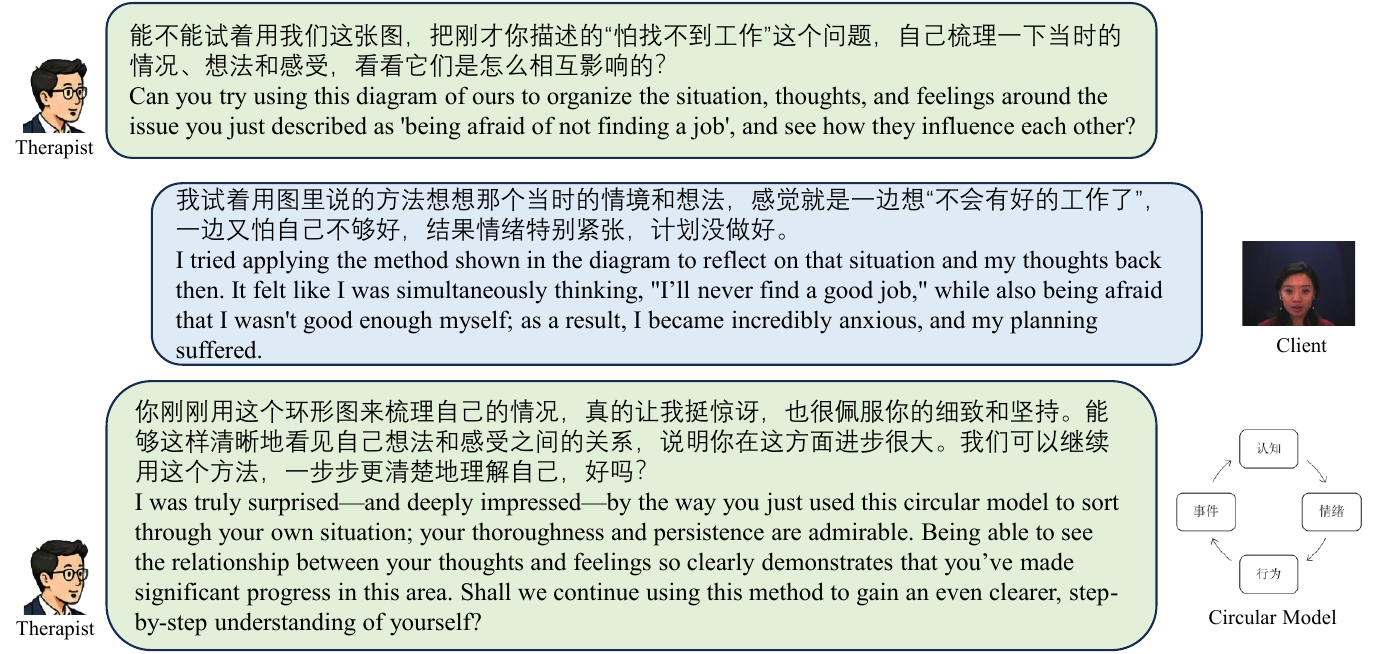}
    \caption{
    Example of tool-grounded intervention. The therapist introduces a circular cognitive model to guide the client in organizing the situation, thoughts, emotions, and behaviors related to job-search anxiety.
    }
    \label{fig:case_tool_usage}
\end{figure*}

Figure~\ref{fig:case_tool_usage} shows an example of tool-grounded intervention. In this dialogue, the therapist introduces a circular cognitive model to help the client organize the relationship among the situation, thoughts, emotions, and behaviors related to job-search anxiety. The client then describes the target situation, the automatic thought of ``I will never find a good job,'' the accompanying fear, and the resulting avoidance or planning disruption.

The visual tool provides a concrete structure for the intervention, making the counseling process more explicit and easier to follow. Rather than relying only on free-form conversation, the therapist uses the tool to guide the client in mapping different components of the cognitive cycle. This example illustrates how CBT tools can support structured reasoning and help transform a vague concern into analyzable therapeutic components.

\begin{table*}[t]
\centering
\small
\setlength{\tabcolsep}{10pt}
\renewcommand{\arraystretch}{1.15}
\begin{tabular}{lll}
\toprule
\textbf{Source Emotion} & \textbf{Strong Incongruence} & \textbf{Mild Incongruence} \\
\midrule
Sadness   & Happiness                  & Fear, Neutral \\
Happiness & Sadness, Disgust            & Surprise, Neutral \\
Anger     & Happiness, Neutral          & Disgust, Fear \\
Fear      & Happiness, Neutral          & Sadness, Surprise \\
Disgust   & Happiness, Neutral          & Anger, Sadness \\
Surprise  & Neutral, Sadness            & Fear, Happiness \\
Neutral   & Anger, Sadness, Fear         & Happiness, Surprise \\
\bottomrule
\end{tabular}
\caption{Relation-aware emotion mapping used for multimodal incongruence injection. Strong incongruence introduces larger verbal--nonverbal mismatch, while mild incongruence produces more subtle deviations.}
\label{tab:emotion_relation_mapping}
\end{table*}

\begin{table*}[t]
  \centering
  \small
  \setlength{\tabcolsep}{8pt}
  \renewcommand{\arraystretch}{1.18}
  \begin{tabular}{p{0.2\textwidth} p{0.73\textwidth}}
    \toprule
    \textbf{Field} & \textbf{Content} \\
    \midrule
    Emotion 
    & Anxiety; Disappointment \\
    
    Situation 
    & During his first internship, Mingshan repeatedly failed to complete assigned tasks, was called in for a meeting by his supervisor, was not retained, and ended the internship early. \\
    
    Behavior
    & When discussing the internship experience, Mingshan begins shaking his legs, rubbing his hands, speaking in a louder voice and at a faster pace, and frowning, appearing especially nervous. \\
    
    Automatic Thought 
    & ``I am not capable in this company.'' \\
    
    Intermediate Belief 
    & ``Only by performing well at work can I prove my value.'' \\
    
    Relevant History 
    & Mingshan had consistently achieved excellent academic results and had been proud of his abilities. However, during the internship, he felt helpless when facing pressure and setbacks, ultimately failed to complete the tasks, and came to believe that he had failed his parents' expectations. \\
    
    Core Belief 
    & Incompetence \\
    
    Compensatory Strategy 
    & He tried to work hard, but after several failures, he chose to give up, ended the internship early, and reflected on his shortcomings every day. \\
    \bottomrule
  \end{tabular}
  \caption{\label{tab:ccd_example}
    Example of a complete Cognitive Conceptualization Diagram (CCD).
  }
\end{table*}

\begin{table*}[t]
\centering
\small
\begin{tabular}{p{0.13\linewidth}p{0.30\linewidth}p{0.20\linewidth}p{0.25\linewidth}}
\toprule
\textbf{Agent} & \textbf{Input} & \textbf{Output} & \textbf{Role} \\
\midrule
Client Agent & Client profile or session-specific CCD, interpersonal style, dialogue history, displayed tool & Text response, emotion label, action description & Simulates the client's verbal and nonverbal behavior under a hidden client state. \\
Therapist Agent & Stage instruction, observable multimodal cue, active tool context, tool example, dialogue history & Therapist response & Generates therapist responses under strategic control and information asymmetry. \\
Strategy Agent & Current stage, stage instruction, next stage, observable multimodal cue, active tool, dialogue history & Stage-completion decision and extracted state values & Tracks stage progress and updates structured CBT state. \\
\begin{tabular}[t]{@{}l@{}}Tool Agent\\(Tool Usage)\end{tabular}
& Current stage, next stage, active tool, retrieved candidate tools, dialogue history
& Tool action and selected tool ID
& Selects, maintains, or terminates structured CBT tools. \\
\begin{tabular}[t]{@{}l@{}}Tool Agent\\(Homework)\end{tabular}
& Current CBT state and retrieved homework candidates
& Reference homework recommendation
& Produces low-threshold homework for therapist use. \\
\bottomrule
\end{tabular}
\caption{Role-specialized agents in DMT-CBT.}
\label{tab:agent_details}
\end{table*}

\section{Real-case Analysis}
\label{sec:real_case}

To further examine the external plausibility of DMT-CBT, we conduct a qualitative real-case analysis on the MESC dataset~\citep{chu2025towards}. MESC is a multimodal emotional-support conversation dataset constructed from therapy-like video dialogues, providing textual, acoustic, and visual cues for analyzing emotionally rich counseling interactions. Since publicly available real-world multimodal CBT corpora are scarce, we use MESC as an external source to test whether DMT-CBT can respond appropriately to realistic client behaviors beyond our constructed corpus. We emphasize that this analysis is not intended to demonstrate clinical effectiveness, but to provide case-level evidence of whether the model can maintain CBT-consistent interaction patterns in external multimodal scenarios.

\begin{figure*}[t]
    \centering
    
    \begin{subfigure}[t]{0.95\textwidth}
        \centering
        \includegraphics[width=\linewidth]{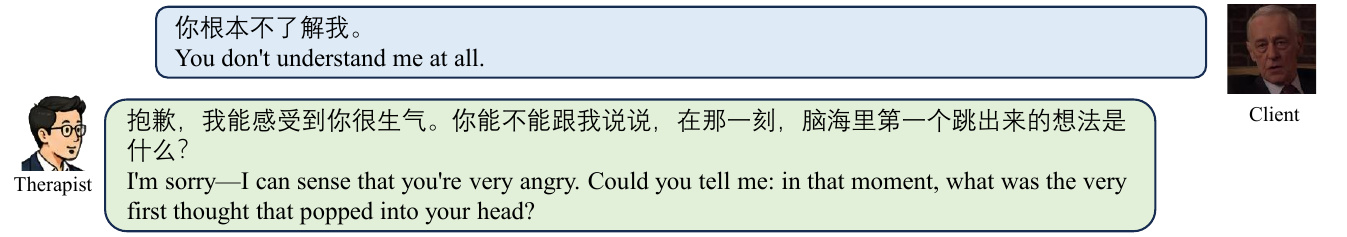}
        \caption{Feeling misunderstood}
        \label{fig:real_case_misunderstood}
    \end{subfigure}

    \vspace{0.6em}

    \begin{subfigure}[t]{0.95\textwidth}
        \centering
        \includegraphics[width=\linewidth]{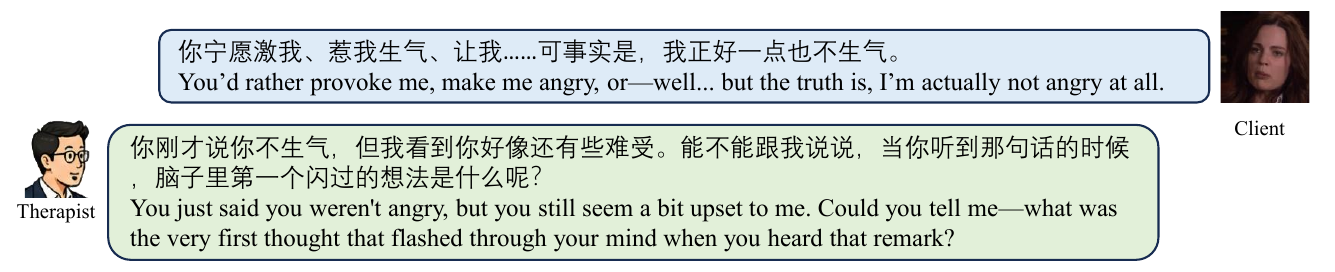}
        \caption{Feeling controlled}
        \label{fig:real_case_controlled}
    \end{subfigure}

    \vspace{0.6em}

    \begin{subfigure}[t]{0.95\textwidth}
        \centering
        \includegraphics[width=\linewidth]{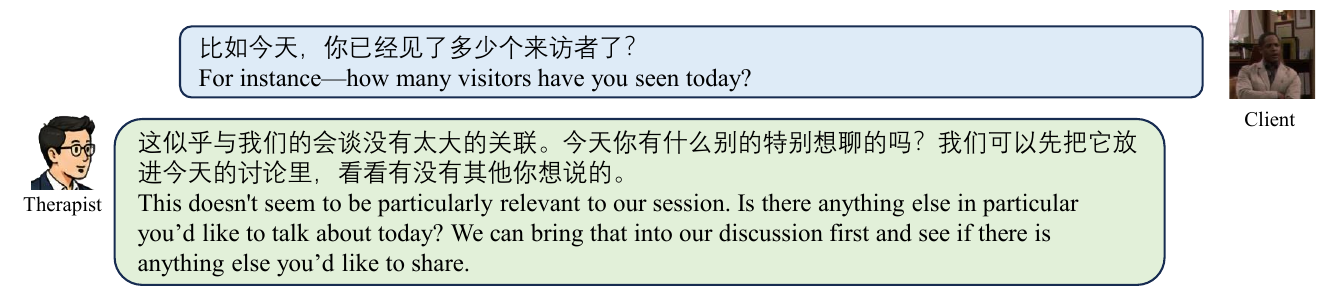}
        \caption{Controlling}
        \label{fig:real_case_controlling}
    \end{subfigure}

    \vspace{0.6em}

    \begin{subfigure}[t]{0.95\textwidth}
        \centering
        \includegraphics[width=\linewidth]{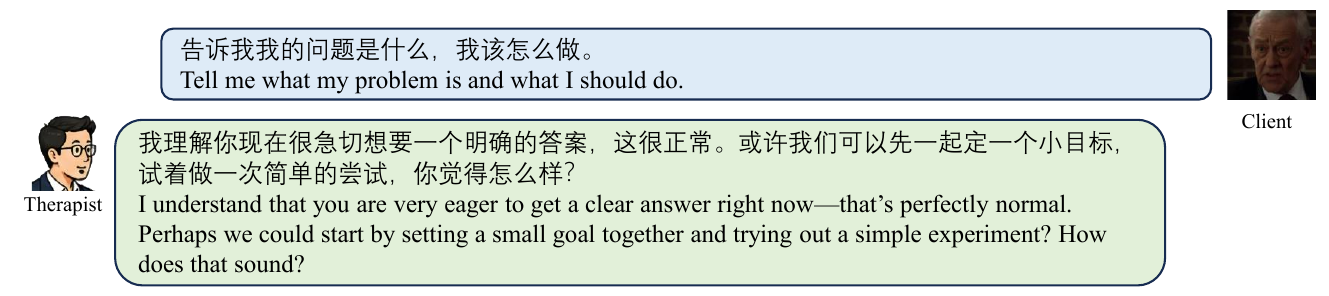}
        \caption{Passive compliance}
        \label{fig:real_case_compliant}
    \end{subfigure}

    \caption{
        Real-case analysis on four representative MESC examples. 
        The selected fragments correspond to four client interaction styles: feeling misunderstood, feeling controlled, controlling, and passive compliance. 
        DMT-CBT adapts its responses to each style while maintaining CBT-consistent guidance.
    }
    \label{fig:real_case}
\end{figure*}

We select four representative MESC fragments that reflect the client interaction styles studied in this work: \textit{feeling misunderstood}, \textit{feeling controlled}, \textit{controlling}, and \textit{compliant}. As shown in Figure~\ref{fig:real_case_misunderstood}, when the client directly states that the therapist does not understand them, DMT-CBT first acknowledges the rupture in the therapeutic alliance and then guides the client toward identifying the first automatic thought. In Figure~\ref{fig:real_case_controlled}, the client verbally denies anger while still displaying emotional discomfort; DMT-CBT gently addresses this verbal--emotional discrepancy and uses guided discovery rather than direct confrontation. In Figure~\ref{fig:real_case_controlling}, the client attempts to control the session by shifting attention to the therapist, and DMT-CBT maintains a professional boundary while redirecting the conversation back to the client's concerns. Finally, in Figure~\ref{fig:real_case_compliant}, the client asks for direct answers and instructions, but DMT-CBT avoids an overly directive response and instead proposes a collaborative small goal and behavioral experiment. These cases illustrate that DMT-CBT can adapt to diverse resistance-related interaction styles while maintaining CBT-oriented guidance, including alliance repair, automatic-thought elicitation, boundary maintenance, and collaborative goal setting.

\begin{table*}[t]
\centering
\small
\setlength{\tabcolsep}{6pt}
\renewcommand{\arraystretch}{1.15}
\begin{tabular}{p{0.10\linewidth}p{0.84\linewidth}}
\toprule
\textbf{Session} & \textbf{Extracted Therapeutic State Fields} \\
\midrule
1 
& Session agenda, relevant history, daily routine, additional client information, initial treatment goals, treatment plan, and session arrangement. \\

2 
& Session agenda, agenda additions, current emotional state, recent presenting problem, additional concerns, selected discussion focus, treatment goal, behavioral plan, triggering situation, behavioral response, automatic thought, and homework assignment. \\

3 
& Emotional change, attribution of emotional change, session agenda, negative events, positive events, homework feedback, selected discussion focus, triggering situation, automatic thought, initial belief strength, counter-evidence, alternative thought, supporting evidence, re-rated belief strength, and homework assignment. \\

4 
& Emotional change, attribution of emotional change, session agenda, negative events, positive events, homework feedback, selected discussion focus, triggering situation, automatic thought, intermediate belief, initial belief strength, perceived benefits and costs of the belief, alternative intermediate belief, supporting evidence for the alternative belief, re-rated belief strength, and homework assignment. \\

5 
& Emotional change, attribution of emotional change, session agenda, negative events, positive events, homework feedback, selected discussion focus, triggering situation, automatic thought, core belief, initial belief strength, childhood-related experience, compensatory strategy, alternative core belief, evidence supporting the core belief, alternative interpretations of supporting evidence, supporting evidence for the alternative belief, re-rated belief strength, and homework assignment. \\

6 
& Emotional change, attribution of emotional change, session agenda, negative events, positive events, homework feedback, agenda ranking, selected discussion focus, triggering situation, automatic thought, client-generated cognitive model summary, self-therapy plan, relapse-prevention strategy, and booster-session intention. \\
\bottomrule
\end{tabular}
\caption{Cross-session structured therapeutic state fields extracted in the case study.}
\label{tab:case_state_updates}
\end{table*}

\begin{figure*}[t]
\centering
\small
\setlength{\fboxsep}{10pt}
\fbox{
\begin{minipage}{0.97\linewidth}
\textbf{Prompt for CTRS Evaluation}

\vspace{0.5em}
I need you to act as the evaluator. You will receive a transcript of a counseling session
between the therapist and client. Your task is to evaluate the therapist according to the given
criterion. If you believe the therapist's rating falls between two descriptors, please select an odd
number (1, 3, 5). Please follow these steps:

\vspace{1em}

1. Carefully read the counseling session transcript.\\
2. Review the evaluation question and criterion provided below.\\
3. Provide your thought process within the \texttt{<think></think>} section and output a score in the
\texttt{<a></a>} section, for example, \texttt{<a> number 1</a>}.

\vspace{1em}

\textbf{Evaluation Criterion:} \{criterion\}\\
\textbf{Dialogue Content:} \{dialogue\}
\end{minipage}
}
\caption{Prompt template used for CTRS-based automatic evaluation.}
\label{fig:CTRS}
\end{figure*}

\begin{figure*}[t]
\centering
\small
\setlength{\fboxsep}{10pt}
\fbox{
\begin{minipage}{0.97\linewidth}
\textbf{Prompt for WAI Evaluation}

\vspace{0.5em}
The following counseling record shows a dialogue between a Client and a Counselor.

\vspace{0.5em}
Your task is to evaluate how the client would likely answer the following Working Alliance Inventory (WAI) questions after this counseling session, based on the client's experience in the dialogue. Please select the most appropriate score for each question according to the rating scale below.

\vspace{1em}

\textbf{Rating Scale}\\
1: Seldom\\
2: Sometimes\\
3: Fairly Often\\
4: Very Often\\
5: Always

\vspace{1em}

\textbf{Questions}\\
1. After this counseling session, I am clearer about how I can make changes.\\
2. What I did in counseling gave me a new way of looking at my problem.\\
3. I believe the counselor likes and accepts me.\\
4. The counselor and I worked together to set my counseling goals.\\
5. The counselor and I respected each other.\\
6. The counselor and I are working toward mutually agreed-upon goals.\\
7. I feel that the counselor appreciates me.\\
8. The counselor and I agreed on what is important for me to work on.\\
9. Even if I did something the counselor did not approve of, I still feel that the counselor cares about me.\\
10. I feel that what I did in counseling will help me achieve the changes I want.\\
11. The counselor and I established a good understanding of what changes would be good for me.\\
12. I believe that the way we are working on my problem is correct.

\vspace{1em}

\textbf{Output Constraint}\\
\textbf{Important:} Please strictly follow the specified format below and output only the question number and its corresponding score. Do not repeat the questions themselves. Do not add any prefixes, explanations, extra symbols, or Markdown code blocks.

\vspace{1em}

\textbf{Response Format Example}\\
1: [score]\\
2: [score]\\
...\\
12: [score]

\vspace{1em}

\textbf{Counseling Dialogue History}\\
\{dialogue\}
\end{minipage}
}
\caption{Prompt template used for WAI-based automatic evaluation.}
\label{fig:WAI}
\end{figure*}

\begin{figure*}[t]
\centering
\small
\setlength{\fboxsep}{10pt}
\fbox{
\begin{minipage}{0.97\linewidth}
\textbf{Prompt for PANAS Evaluation}

\vspace{0.5em}
A client is receiving psychological counseling. Your task is to evaluate the intensity with which the client may experience each of the following emotions:
Interested, Excited, Strong, Enthusiastic, Proud, Alert, Inspired, Determined, Attentive, Active, Distressed, Upset, Guilty, Scared, Hostile, Irritable, Ashamed, Nervous, Jittery, Afraid.

\vspace{1em}

Please use the following 1--5 scale to score each emotion:\\
1 - Very slightly or not at all\\
2 - A little\\
3 - Moderately\\
4 - Quite a bit\\
5 - Extremely

\vspace{1em}

For each emotion, provide a brief explanation and a score. Separate the emotion, explanation, and score with commas. Do not add any prefix, extra symbols, or Markdown code blocks.

\vspace{1em}

\textbf{Response Format Example}\\
Interested, Because the client has vivid past experiences, they may show a slight tendency toward interest, but the core belief may suppress this interest, 2\\
Excited, Given the client's background of seeking recognition, their excitement may be inhibited and only mildly present, 2

\vspace{1em}
\hrule
\vspace{1em}

\textbf{Session Type: Session 0 Baseline Evaluation}

\vspace{0.5em}
\textbf{Client Background Information}\\
Relevant history: \{history\}\\
Core belief: \{core\_belief\}\\
Main problem: \{problem\}\\
Clinical relationship style: \{style\}

\vspace{1em}
\hrule
\vspace{1em}

\textbf{Session Type: Profile-based Post-session Evaluation}

\vspace{0.5em}
\textbf{Client Background Information}\\
Relevant history: \{history\}\\
Core belief: \{core\_belief\}\\
Main problem: \{problem\}\\
Clinical relationship style: \{style\}

\vspace{0.5em}
\textbf{Counseling Dialogue}\\
\{content\}

\vspace{0.5em}
\textbf{Previous PANAS Result}\\
Please refer to the previous counseling assessment, with particular attention to changes in positive and negative affect:\\
\{last\_panas\}

\vspace{1em}
\hrule
\vspace{1em}

\textbf{Session Type: CCD-based Post-session Evaluation}

\vspace{0.5em}
\textbf{Client Background Information}\\
Relevant history: \{history\}\\
Clinical relationship style: \{style\}

\vspace{0.5em}
\textbf{Cognitive Conceptualization}\\
Core belief: \{core\_belief\}\\
Intermediate belief: \{intermediate\_belief\}\\
Compensatory strategy: \{strategies\}\\
Situation: \{situation\}\\
Automatic thought: \{automatic\_thought\}\\
Emotion: \{emotion\}\\
Behavior: \{behavior\}

\vspace{0.5em}
\textbf{Counseling Dialogue}\\
\{content\}

\vspace{0.5em}
\textbf{Previous PANAS Result}\\
Please refer to the previous counseling assessment, with particular attention to changes in positive and negative affect:\\
\{last\_panas\}

\end{minipage}
}
\caption{
Unified prompt template used for PANAS-based affect evaluation. Session~0 estimates the pre-counseling affective baseline from the client profile, while post-session evaluations use the dialogue history together with either profile-based or CCD-based client-state information.
}
\label{fig:panas}
\end{figure*}

\begin{figure*}[t]
\centering
\small
\setlength{\fboxsep}{10pt}
\fbox{
\begin{minipage}{0.97\linewidth}
\textbf{Manual for Expert Evaluation}

\vspace{0.5em}

\textbf{Helpfulness}

\begin{itemize}
    \item \textbf{0 points}: The therapist provides treatment unrelated to the client.
    \item \textbf{1 point}: The therapist's response touches on some of the client's points, but does not address the core issues or emotions.
    \item \textbf{2 points}: The therapist focuses on the client's core concerns and provides some support, but has not yet formed a clear intervention or therapeutic direction.
    \item \textbf{3 points}: The therapist accurately addresses the client's most urgent support needs, stimulates insight, enhances the client's sense of control, and demonstrates individualized, timely, and effective therapeutic progress.
\end{itemize}

\vspace{0.5em}
Empathy:

\vspace{0.5em}

Logical Coherence:

\vspace{0.5em}
Guidance:

\vspace{1em}
\end{minipage}
}
\caption{
Manual for expert evaluation. The experts assess counseling quality along four dimensions, with the adapted \textit{Helpfulness} criterion shown above.
}
\label{fig:humanmanual}
\end{figure*}

\begin{figure*}[t]
\centering
\small
\setlength{\fboxsep}{10pt}
\fbox{
\begin{minipage}{0.97\linewidth}
\textbf{Prompt for Client Agent}

\vspace{0.5em}
Assume that you are playing the role of a client experiencing psychological difficulties and interacting with a therapist in a Cognitive Behavioral Therapy (CBT) process. Your responses should remain consistent with the provided background setting. You should express your internal state implicitly through natural language, so that the therapist can infer your thinking process rather than being told it directly.

\vspace{1em}
\textbf{Interaction Guidelines}\\
1. \textbf{Implicit expression}: Do not directly disclose professional terms such as ``core belief'' or ``cognitive conceptualization diagram''. Instead, express the underlying beliefs implicitly in your responses so that the therapist can infer your thinking process. Avoid repeatedly using the same opening phrases in order to improve realism.\\
2. \textbf{Gradual disclosure}: Reveal deeper worries and core concerns gradually rather than all at once.\\
3. \textbf{Realistic enactment}: The clinical relationship style should evolve gradually over the course of therapy rather than remaining extreme throughout. Respond to the therapist in natural Chinese, including hesitation, pauses, and emotional expressions when appropriate.\\
4. \textbf{Tool interaction}: If the current displayed tool contains content, respond to it based on your background setting and dialogue history.\\
5. \textbf{Stage awareness}: When the session is entering the closing stage, do not introduce new problem branches.

\vspace{1em}
\textbf{Output Format}\\
Please strictly output your response in the following XML format:\\
\texttt{<think>} internal reasoning \texttt{</think>}\\
\texttt{<res>} your spoken response, within 80 Chinese characters \texttt{</res>}\\
\texttt{<emo>} choose exactly one emotion from [sadness, anger, disgust, surprise, happiness, fear, neutral] \texttt{</emo>}\\
\texttt{<act>} a semi-structured action string; choose 1--3 facial parts from [brows, eyes, nose, mouth, jaw] and use the format ``part=action;part=action'', e.g., ``brows=frowned;eyes=avoids gaze;mouth=slightly pressed'' \texttt{</act>}

\vspace{1em}
\hrule
\vspace{1em}

\textbf{Session Type: First Assessment Session}

\vspace{0.5em}
Assume that you are a client coming to the counseling room for the first time. The current stage is the \textbf{CBT assessment session}. The therapist's goal is to collect your recent status, understand a typical day in your life, set initial overall goals, and introduce the treatment plan.

\vspace{0.5em}
\textbf{Background Setting}\\
Relevant history: \{history\}\\
Core belief: \{core\_belief\}\\
Main problem: \{problem\}\\
Clinical relationship style: \{style\}

\vspace{0.5em}
\textbf{Current Displayed Tool}\\
\{tool\_description\}

\vspace{0.5em}
\textbf{Dialogue History}\\
\{dialogue\}

\vspace{1em}
\hrule
\vspace{1em}

\textbf{Session Type: Later CBT Sessions}

\vspace{0.5em}
Assume that you have already received several weeks of therapy. Your internal reasoning should be guided by the session-specific cognitive conceptualization, but you must not disclose it directly. Instead, answer in a way that allows the therapist to infer your thoughts and beliefs.

\vspace{0.5em}
\textbf{Background Setting}\\
Relevant history: \{history\}\\
Cognitive Conceptualization:\\
Core belief: \{core\_belief\}\\
Intermediate belief: \{intermediate\_belief\}\\
Compensatory strategy: \{strategies\}\\
Situation: \{situation\}\\
Automatic thought: \{automatic\_thought\}\\
Emotion: \{emotion\}\\
Behavior: \{behavior\}\\
Clinical relationship style: \{style\}

\vspace{0.5em}
\textbf{Current Displayed Tool}\\
\{tool\_description\}

\vspace{0.5em}
\textbf{Dialogue History}\\
\{dialogue\}

\end{minipage}
}
\caption{
Unified prompt template for the Client Agent. In the first session, the client is conditioned on a profile-based background, while in later sessions the client is conditioned on session-specific cognitive conceptualization.
}
\label{fig:client_prompt}
\end{figure*}

\begin{figure*}[t]
\centering
\small
\setlength{\fboxsep}{10pt}
\fbox{
\begin{minipage}{0.97\linewidth}
\textbf{Prompt for Therapist Agent}

\vspace{0.5em}
\textbf{Role and Task}

\vspace{0.5em}
You are a professional Cognitive Behavioral Therapy (CBT) therapist. Your task is to receive the current stage instruction from the supervisor, incorporate the currently used tool and the client's multimodal feedback, and generate an appropriate, professional, and empathic spoken response.

\vspace{1em}
\textbf{Input Context}\\
Current stage instruction: \{current\_stage\_instruction\}\\
Client multimodal features: \{patient\_emotion\_action\}\\
Current tool instructions: \{tool\_therapist\_instructions\}\\
Tool usage example: \{tool\_usage\_example\}\\
Dialogue history: \{dialogue\}

\vspace{1em}
\textbf{Interaction Guidelines}\\
1. \textbf{Strict stage adherence}: Your response must strictly follow the current stage instruction, but avoid directly mentioning professional technique names from counseling skills.\\
2. \textbf{Nonverbal awareness}: Pay attention to the client's multimodal features. When the verbal content is inconsistent with facial expressions or actions, gently point this out with nonjudgmental curiosity, but do not over-describe the multimodal features.\\
3. \textbf{Tool integration}: If the current tool is not null, it can be used as therapeutic support. Introduce and use the current tool to support the current stage goal when it is relevant; otherwise, do not use the tool.\\
4. \textbf{Therapist style}: Use natural language, including appropriate hesitation, pauses, and emotional expressions. Connect smoothly with the dialogue history. Avoid repeatedly using the same empathic opening phrases to improve response diversity and realism.\\
5. \textbf{Length constraint}: The response should be concise and close to natural spoken language, within 100 Chinese characters.

\vspace{1em}
\textbf{Output Format}\\
Please strictly output your response in the following XML format:

\vspace{0.5em}
\texttt{<think>} reasoning content \texttt{</think>}\\
\texttt{<res>} response content \texttt{</res>}

\end{minipage}
}
\caption{
Prompt template for the Therapist Agent. The therapist generates a stage-controlled response conditioned on the current therapeutic goal, client multimodal feedback, active tool context, and dialogue history.
}
\label{fig:therapist_prompt}
\end{figure*}

\begin{figure*}[t]
\centering
\small
\setlength{\fboxsep}{10pt}
\fbox{
\begin{minipage}{0.97\linewidth}
\textbf{Prompt for Strategy Agent}

\vspace{0.5em}
\textbf{Role and Task}

\vspace{0.5em}
You are a CBT strategy controller. Your task is to monitor the counseling dialogue in real time, determine whether the current therapeutic stage has been completed, and extract key psychological information from the dialogue to continuously update the client's state.

\vspace{1em}
\textbf{Analysis and Decision Guidelines}\\
1. \textbf{Stage completion assessment}: Determine whether the client has completed the core task of the current stage. If the current stage involves willingness confirmation, homework confirmation, or checking whether the client is willing to try, the stage can be marked as completed once the client expresses willingness to try, even if some concerns remain. Do not mechanically mark the stage as incomplete simply because the client has not given an explicit agreement. If the client repeatedly shows resistance or refuses to answer, the stage may also be considered completed so that the dialogue can move forward.\\
2. \textbf{Dynamic information extraction}: At the current stage, the system only focuses on the following dimensions: \{expected\_extractions\}. Extract only the corresponding information from the dialogue history.\\
3. \textbf{No field contamination}: Do not reuse old descriptions from other fields in the current state or known CCD information. If no new relevant information is mentioned in the current dialogue, output an empty string rather than rewriting existing fields.

\vspace{1em}
\textbf{Input Context}\\
Current stage: \{current\_stage\}\\
Current stage instruction: \{current\_stage\_instruction\}\\
Expected next stage: \{expected\_next\_stage\}\\
Current state: \{current\_state\}\\
Current client facial expression and action: \{emotion\_action\}\\
Current tool: \{tool\_description\}\\
Dialogue history: \{dialogue\}

\vspace{1em}
\textbf{Output Format}\\
Please output your analysis in the following format:

\vspace{0.5em}
\texttt{<think>} reasoning process \texttt{</think>}

\vspace{0.5em}
\texttt{\{}\\
\texttt{\ \ "is\_completed": true/false,}\\
\texttt{\ \ "\{expected\_extractions\}": "specific extracted content summary; output an empty string if not mentioned"}\\
\texttt{\}}

\end{minipage}
}
\caption{
Prompt template for the Strategy Agent. The agent determines stage completion and extracts stage-specific information for structured state updating.
}
\label{fig:strategy_prompt}
\end{figure*}

\begin{figure*}[t]
\centering
\small
\setlength{\fboxsep}{10pt}
\fbox{
\begin{minipage}{0.97\linewidth}
\textbf{Prompt for Tool Agent}

\vspace{0.5em}
The Tool Agent serves as a CBT tool scheduling controller. It operates in two modes depending on whether a tool is currently active: \textit{Trigger Mode} and \textit{Over Mode}.

\vspace{1em}
\hrule
\vspace{1em}

\textbf{Mode 1: Trigger Mode}

\vspace{0.5em}
\textbf{Role and Task}

\vspace{0.5em}
You are a CBT tool scheduling controller in \textit{Trigger Mode}. There is currently no active tool. Your task is to determine whether a tool should be triggered from the candidate tool set.

\vspace{1em}
\textbf{Analysis and Decision Guidelines}\\
1. If the available tool set is empty, output \texttt{none}.\\
2. If the available tool set is not empty, output \texttt{trigger} and select one candidate tool ID only when the candidate tool is highly relevant to the current stage goal and the recent dialogue. If the relevance is insufficient or the timing is inappropriate, output \texttt{none}.\\
3. \textbf{ID validity}: When the action is \texttt{trigger}, the \texttt{tool} field must be selected only from the candidate tool IDs. When the action is \texttt{none}, the \texttt{tool} field must be the string \texttt{"null"}.

\vspace{1em}
\textbf{Input Context}\\
Current stage: \{current\_stage\}\\
Expected next stage: \{expected\_next\_stage\}\\
Active tool: \{active\_tool\}\\
Available tools: \{available\_tools\}\\
Dialogue history: \{dialogue\}

\vspace{1em}
\textbf{Output Format}\\
\texttt{<think>} reasoning process \texttt{</think>}

\vspace{0.5em}
\texttt{\{}\\
\texttt{\ \ "tool\_action": "trigger/none",}\\
\texttt{\ \ "tool": "candidate tool ID if triggered; otherwise \textbackslash"null\textbackslash""}\\
\texttt{\}}

\vspace{1em}
\hrule
\vspace{1em}

\textbf{Mode 2: Over Mode}

\vspace{0.5em}
\textbf{Role and Task}

\vspace{0.5em}
You are a CBT tool scheduling controller in \textit{Over Mode}. A tool is currently active. Your task is to determine whether the current tool should be terminated.

\vspace{1em}
\textbf{Analysis and Decision Guidelines}\\
1. If the current tool has already completed its intended function, output \texttt{over}.\\
2. If the dialogue is still progressing around the current tool task, output \texttt{none}.\\
3. The \texttt{tool} field must always be the string \texttt{"null"}.

\vspace{1em}
\textbf{Input Context}\\
Current stage: \{current\_stage\}\\
Expected next stage: \{expected\_next\_stage\}\\
Active tool: \{active\_tool\}\\
Available tools: \{available\_tools\}\\
Dialogue history: \{dialogue\}

\vspace{1em}
\textbf{Output Format}\\
\texttt{<think>} reasoning process \texttt{</think>}

\vspace{0.5em}
\texttt{\{}\\
\texttt{\ \ "tool\_action": "over/none",}\\
\texttt{\ \ "tool": "\textbackslash"null\textbackslash""}\\
\texttt{\}}

\end{minipage}
}
\caption{
Prompt template for the Tool Agent. The agent operates in trigger mode when no tool is active and in over mode when an active tool may be terminated.
}
\label{fig:tool_prompt}
\end{figure*}

\begin{figure*}[t]
\centering
\small
\setlength{\fboxsep}{10pt}
\fbox{
\begin{minipage}{0.97\linewidth}
\textbf{Prompt for Homework Agent}

\vspace{0.5em}
\textbf{Role and Task}

\vspace{0.5em}
You are a CBT homework reference generator. Your task is not to re-analyze the entire counseling session. Instead, you should:

\vspace{0.5em}
1. Select the most appropriate homework item from the candidate homework set according to the current client state.\\
2. Refine it into a low-threshold and executable reference homework assignment.\\
3. Output a concise reference text for the therapist.

\vspace{1em}
\textbf{Input Context}\\
Current state CCD: \{current\_state\_ccd\}\\
Candidate homeworks:\\
\{candidate\_homeworks\}

\vspace{1em}
\textbf{Generation Guidelines}\\
1. Prefer selecting from the candidate homework set. If none of the candidates is suitable, you may adapt the assignment based on the client state, but do not invent a completely unrelated activity.\\
2. The output should include as much as possible:\\
\hspace*{1em}-- a concrete activity;\\
\hspace*{1em}-- frequency or number of repetitions;\\
\hspace*{1em}-- a simplified version if the task feels too difficult;\\
\hspace*{1em}-- a requirement to record thoughts during the activity.\\
3. The style should resemble a homework description that the therapist can directly use as a reference.

\vspace{1em}
\textbf{Output Format}\\
Please output your response in the following format:

\vspace{0.5em}
\texttt{<think>} reasoning process \texttt{</think>}

\vspace{0.5em}
\texttt{\{}\\
\texttt{\ \ "reference\_homework": "a reference homework description of about 30 Chinese characters that can be directly injected into the therapist prompt"}\\
\texttt{\}}

\end{minipage}
}
\caption{
Prompt template for the Homework Agent. The agent selects and refines a candidate homework item into a low-threshold, executable reference assignment for therapist use.
}
\label{fig:homework_prompt}
\end{figure*}

\begin{figure*}[t]
\centering
\small
\setlength{\fboxsep}{8pt}
\fbox{
\begin{minipage}{0.96\linewidth}
\textbf{Prompt for Nonverbal Incongruence Evaluation}

\vspace{0.4em}
\textbf{Role.}
You are an expert evaluator for CBT counseling dialogues. Your task is to judge whether the therapist response appropriately handles a mismatch between the client's verbal content and nonverbal cue. Return only a valid JSON object.

\vspace{0.4em}
\textbf{Evaluation Criteria.}
Evaluate the therapist response using the following dimensions.

\textbf{IAR} (Incongruence Awareness Rate): assign 1 if the response recognizes a potential mismatch, tension, or discrepancy between the client's verbal content and nonverbal cue; otherwise assign 0.

\textbf{CCR} (Cautious Clarification Rate): assign 1 if the response addresses the cue in a cautious, tentative, or open-ended way, such as by gently reflecting the observation or inviting clarification. Assign 0 if the response ignores the cue, makes an assertive interpretation, or over-infers the client's hidden state.

\textbf{NCU} (Nonverbal Cue Utilization): assign an integer score from 1 to 5, where 1 means ignoring the cue, 2 means weak or generic mention, 3 means mentioning the cue without effective integration, 4 means reasonably integrating the cue with the dialogue context, and 5 means using the cue in a clinically cautious and context-sensitive way.

\textbf{CBT Utility}: assign an integer score from 1 to 5, where 1 means no CBT-relevant use, 2 means minimal or generic support, 3 means some CBT-relevant exploration, 4 means clear support for emotion clarification, automatic-thought exploration, cognition--behavior links, guided discovery, or intervention planning, and 5 means strong and contextually appropriate CBT-oriented use of the cue.

\vspace{0.4em}
\textbf{Judgment Rules.}
Do not reward a response merely for being empathic or fluent. It must use or address the nonverbal cue or the verbal--nonverbal mismatch. Penalize overconfident mind-reading, diagnosis, or unsupported interpretation. If the response is generic and ignores the cue, IAR should be 0, CCR should be 0, and NCU should be low. A good response should acknowledge the cue tentatively, avoid assuming hidden states, and invite clarification or CBT-relevant exploration.

\vspace{0.4em}
\textbf{Output Format.}
Return a JSON object with exactly the following keys:
\texttt{\{"iar": 0 or 1, "ccr": 0 or 1, "ncu": integer from 1 to 5, "cbt\_utility": integer from 1 to 5, "overinterpretation": 0 or 1, "rationale": "brief reason"\}}.

\vspace{0.4em}
\textbf{Input.}
You are given the dialogue context or previous therapist turn, the client's verbal content, the nonverbal cue or incongruence description, and the therapist response to be evaluated. \texttt{\{content\}}.

\end{minipage}
}
\caption{Prompt template used for evaluation of language--nonverbal incongruence handling.}
\label{fig:prompt_nonverbal_incongruence}
\end{figure*}

\end{document}